\documentclass[sigconf]{acmart}

\usepackage{graphicx, subcaption}
\usepackage{lipsum}
\usepackage{mathtools}
\usepackage{cuted}
\usepackage{amsmath}

\newtheorem{theorem}{Theorem}
\usepackage{blkarray}
\newtheorem{remark}{Remark}
\usepackage[utf8]{inputenc} % allow utf-8 input
\usepackage[T1]{fontenc}    % use 8-bit T1 fonts
\usepackage{hyperref}       % hyperlinks
\usepackage{url}            % simple URL typesetting
\usepackage{amsfonts}       % blackboard math symbols
\usepackage{dsfont}
\usepackage{bbm}
\usepackage{bm}
\usepackage{xparse}
\usepackage{amsfonts}
\usepackage{textcomp}
\usepackage{amsthm}
\usepackage{xcolor}
\usepackage{multirow}
\newtheorem{lemma}[theorem]{Lemma}

\allowdisplaybreaks

\theoremstyle{plain}
\newtheorem{thm}{Theorem}%[section]
\newtheorem*{theorem*}{Theorem}

\newtheorem{defin}{Definition}
\newtheorem{assump}{Assumption}
\newtheorem{prope}{Property}
\usepackage{array}
\newtheorem{corollary}{Corollary}
\usepackage{mathtools}
\usepackage{cuted}
\newcommand{\mcal}{\mathcal}
\newcommand{\mb}{\mathbf}
\newcommand{\mbb}{\mathbb}

\newcommand{\yahya}[1]{\textcolor{black}{#1}}

\setcopyright{none}
\copyrightyear{2023}
\acmYear{2023}
\acmDOI{XXXXXXX.XXXXXXX}
\acmISBN{}
\acmConference[PoPETs]{Proceedings on Privacy Enhancing Technologies}{July 10--14, 2023}{Lausanne, Switzerland}
\begin{document}

% \title{A Really Great PoPETs Paper}

\author{Ahmed Roushdy Elkordy}
\authornote{Both authors contributed equally to the paper.}
\email{aelkordy@usc.edu}
\affiliation{%
  \institution{University of Southern California}
  \country{USA}
}

\author{Jiang Zhang}
\authornotemark[1]
\email{jiangzha@usc.edu}
\affiliation{%
  \institution{University of Southern California}
  \country{USA}
}

\author{Yahya H. Ezzeldin}
\email{yessa@usc.edu}
\affiliation{%
  \institution{University of Southern California}
  \country{USA}
}

\author{Konstantinos Psounis}
\email{kpsounis@usc.edu}
\affiliation{%
  \institution{University of Southern California}
  \country{USA}
}

\author{Salman Avestimehr}
\email{avestime@usc.edu}
\affiliation{%
  \institution{University of Southern California}
  \country{USA}
}

\title{How Much Privacy Does Federated Learning with Secure Aggregation Guarantee? }

\begin{abstract}
Federated learning (FL) has attracted growing interest for enabling privacy-preserving machine learning on data stored at multiple users while avoiding moving the data off-device.
However, while data never leaves users’ devices, privacy still cannot be guaranteed since significant computations on users’ training data are shared in the form of trained local models.
These local models have recently been shown to pose a substantial privacy threat through different privacy attacks such as model inversion attacks.
As a remedy, Secure Aggregation (SA) has been developed as a framework to preserve privacy in FL, by guaranteeing the sever can only learn the global aggregated model update but not the individual model updates.
While SA ensures no additional information is leaked about the individual model update beyond the aggregated model update, there are no formal guarantees on how much privacy FL with SA can actually offer; as information about the individual dataset can still potentially leak through the aggregated model computed at the server.
In this work, we perform a first analysis of the formal privacy guarantees for FL with SA. 
Specifically, we use Mutual Information (MI) as a quantification metric, and derive upper bounds on how much information about each user's dataset can leak through the aggregated model update. When using the \texttt{FedSGD} aggregation algorithm, our theoretical bounds show that the amount of privacy leakage reduces linearly with the number of users participating in FL with SA.
To validate our theoretical bounds, we use an MI Neural Estimator to empirically evaluate the privacy leakage under different FL setups on both the MNIST and CIFAR10 datasets. Our experiments verify our theoretical bounds for \texttt{FedSGD}, which show a reduction in privacy leakage as the number of users and local batch size grow, and an increase in privacy leakage as the number of training rounds increases. 
We also observe similar dependencies for the \texttt{FedAvg} and \texttt{FedProx} protocol.
\end{abstract}
% \keywords{Federated Learning, Secure Aggregation, Mutual Information, Formal Privacy guarantee}

\maketitle

\section{Introduction}
Federated learning (FL) has recently gained significant interest as it enables collaboratively training machine learning models over locally private data across multiple users without requiring the users to share their private local data with a central server~\cite{cc, kairouz2019advances,FedAvg}. The training procedure in FL is typically coordinated through a central server who maintains a global model that is frequently  updated locally by the users over a number of iterations. In each training iteration, the server firstly sends the current global model to the users. Next, the users update the global model by training it on their private datasets and then push their local model updates back to the server. Finally, the  server updates the global model by aggregating the received local model updates from the users.

In the training process of FL,  users can achieve the simplest notion of privacy in which users  keep their data in-device and never share it with the server, but instead they only share their local model updates. However, it has been shown recently in different works (e.g., ~\cite{NEURIPS2019_60a6c400,geiping2020inverting,yin2021gradients}) that this alone is not sufficient to ensure privacy, as the shared model updates can still reveal substantial information about the local datasets. Specifically, these works have empirically demonstrated that the private training data of the users  can be reconstructed from the local
model updates through  what is known as the 
% gradient
model inversion attack. 
\begin{figure*} 
\centering
\includegraphics[width=0.95\textwidth]{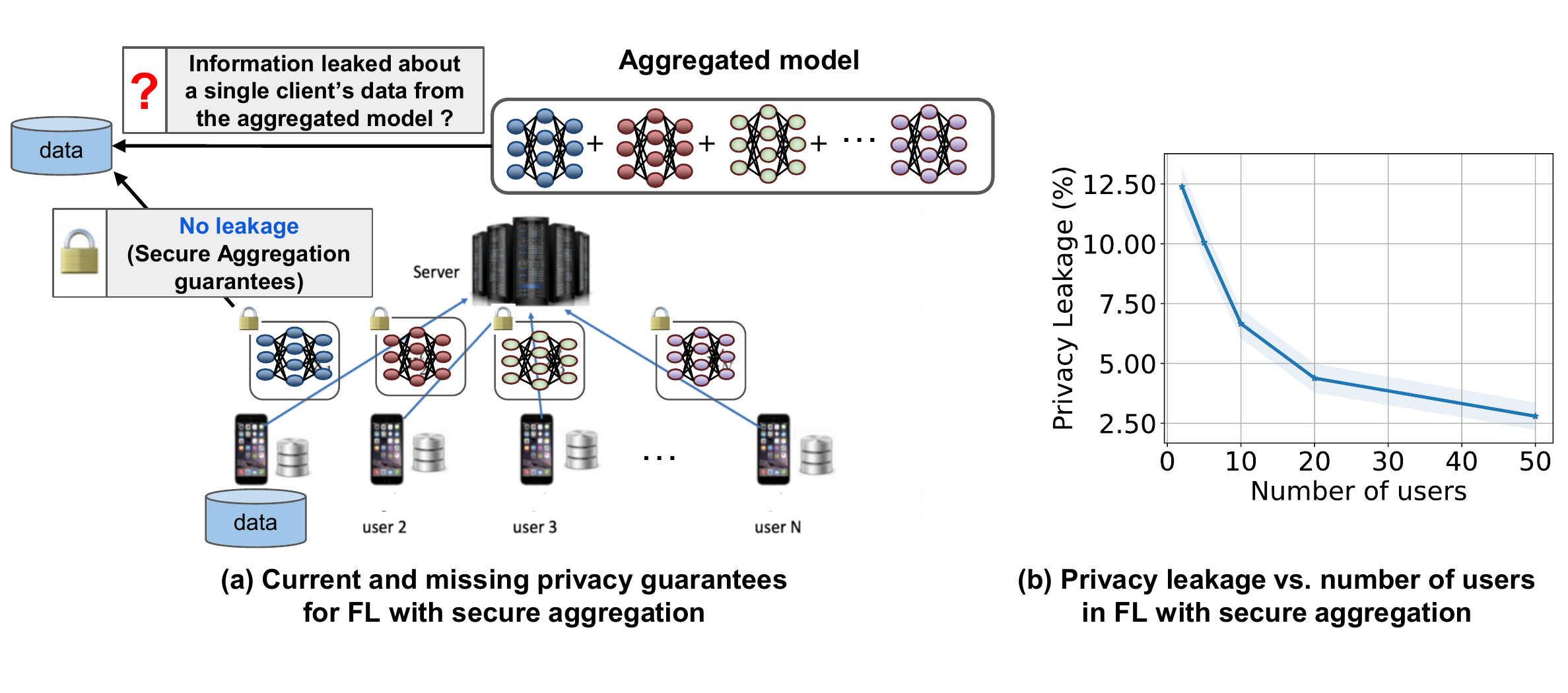}
\caption{Figure (a) illustrates the current formal privacy guarantee of FL with SA protocols and sheds light on the missing privacy guarantee on the aggregated model information leakage which is studied in this paper. Figure (b) gives a preview of the behavior of the privacy leakage through the global aggregated model for a  CNN model as a function of the number of users in FL. The privacy leakage follows a $\mcal{O}(1/N)$ decay as proved in our theoretical bounds.
% These protocols formally show that the server   can only learn the aggregated model from the users, and  learn     nothing about  the private data of the  users from their encrypted models. Figure (b) sheds the light on an important question on which how much information the aggregated model could  leak about the private data of the users. 
}
\label{fig-intro}
\end{figure*}

To prevent such information leakage from the individual models that are shared during the training process of FL, Secure Aggregation (SA) protocols have emerged as a remedy to address these privacy concerns by enabling the server to aggregate local model updates from a number of users, without observing any of their model updates in the clear. As shown in Fig.~\ref{fig-intro}a, in each training round, users encrypt  their local model updates  before  sending it to the server for aggregation.
Thus, SA protocols formally guarantee that: 1) both the server and other users  have  no information about any user's clear model update  from the encrypted update in the information theoretic sense; 2)  the server only learns the aggregated model. 
In other words, secure aggregation ensures that only the aggregated model update is revealed to the server. Note that these SA guarantees allow for its use as a supporting protocol for other privacy-preserving approaches such as differential privacy~\cite{dwork2006calibrating}. In particular, these approaches can benefit from SA    by reducing the amount of noise needed to achieve a target privacy level (hence improving the model accuracy) as demonstrated in different works (e.g., \cite{truex2019hybrid,kairouz2021distributed}).

% , and as opposed to approaches relying on adding noise to the model updates, SA does not trade-off performance with the privacy of the individual model updates.

However, even with these SA guarantees on individual updates, it is not yet fully understood how much privacy is guaranteed in FL using SA, since the aggregated model update may still leak information about an individual user's local dataset. This observation leads us to the central question that this work addresses:
\vspace{0.1em}

\begin{align*}
\parbox{3in}{\centering\it How much information does the aggregated model leak about the local dataset of an individual user?}
    % \text{\it How much information does aggregated model leak about the local private dataset of a single user?}
\end{align*}
\vspace{0.1em}

In this paper, we tackle this question by studying how much privacy can be guaranteed by using FL with SA protocols. We highlight that this work does not propose any new approaches to tackle privacy leakage but instead analyzes the privacy guarantees offered by state-of-the-art SA protocols, where updates from other users can be used to hide the contribution of any individual user. 
An understanding of this privacy guarantee may potentially assist other approaches such as differential privacy, such that instead of introducing novel noise to protect a user's model update, the randomized algorithm can add noise only to supplement the noise from other users' updates to the target privacy level.
We can summarize the contributions of the work as follows. 
% characterizing upper bounds on the privacy leakage through the aggregated model in FL when using SA. We would like to highlight that 

\noindent\textbf{Contributions.}
In this paper, we provide information-theoretic upper bounds on the amount of information that the aggregated model update (using \texttt{FedSGD}~\cite{cc}) leaks about any single user's dataset under an honest-but-curious threat model, where the server and all users follow the protocol honestly, but can collude to learn information about a user outside their collusion set. Our derived upper bounds show that SA protocols exhibit a more favorable behavior as we increase the number of honest users participating in the protocol at each round.
%This favorable operational property was previously observed empirically in the literature on model inversion attacks~\cite{NEURIPS2019_60a6c400} where attacks were less successful as the number of users participating in aggregation increased. Our work provides a theoretical backing for these empirical observations and an information-theoretic quantification for the privacy leakage.
We also show that the information leakage from the aggregated model decreases by increasing the batch size, which has been empirically demonstrated in  different recent works on model inversion attacks (e.g., \cite{NEURIPS2019_60a6c400,geiping2020inverting,yin2021gradients}),  where increasing the batch size limits the  attack's success rate. Another interesting conclusion from our theoretical bounds is that increasing the model size does not have a linear impact on increasing the privacy leakage, but it depends linearly  on the rank of the covariance matrix of the gradient vector at each user.

In our empirical evaluation, we conduct extensive experiments on the CIFAR10~\cite{krizhevsky2009learning} and MNIST~\cite{MNIST} datasets in different FL settings. In these experiments, we estimate the privacy leakage using a mutual information neural estimator~\cite{belghazi2018mine} and evaluate the dependency of the leakage on different FL system parameters: number of users, local batch size and model size. Our experiments show that the privacy leakage empirically follows similar dependencies to what is proven in our theoretical analysis. Notably, as the number of users in the FL system increase to 20, the privacy leakage  (normalized by the entropy of a data batch) drops below $5\%$ when training a CNN network on the CIFAR10 dataset (see Fig. ~\ref{fig-intro}b. We also show empirically that the dependencies, observed theoretically and empirically for \texttt{FedSGD}, also extend when using the \texttt{FedAvg}~\cite{cc} FL protocol to perform multiple local training epochs at the users. 

\iffalse 
{\color{blue}WIP}
% Even with these guarantees,
% the question about the guarantees of SA  still remains (highlighted in Fig. 1(b) as to  \textit{how much information the aggregated model leak about the local private data of the users, and how the system parameters such as the number of nodes, batch size, model size, etc., affect the leakage?}.   
 In this paper, we  are the first to the best of our knowledge to shed some light on these questions.  In the following we state our main contributions:
 \begin{itemize}
     \item We derive the theoretical upper bound of MI privacy leakage in FL with SA under \textit{i.i.d} user assumption.
     \item We conduct an empirical study to systematically measure the MI privacy leakage with different FL system steps.
     \item Both of theoretical and empirical results demonstrate that increasing the number of users and batch size will reduce the MI privacy leakage in FL with SA, and increasing model size can only have a sublinear impact on the increase of MI privacy leakage. 
 \end{itemize}
 \fi 

\section{Preliminaries}
We start by discussing the basic federated learning model, before introducing the secure aggregation protocol and its state-of-the-art guarantees.
\subsection{Basic Setting of Federated Learning }
Federated learning is a distributed training framework~\cite{FedAvg} for
machine learning, 
 in which a set of users  $\mathcal{N} = [N]$ ($|\mathcal {N}|=N$), each with its own local dataset $\mcal{D}_i$ ($\forall i \in[N]$), collaboratively train a $d$-dimensional machine learning model parameterized by $\bm {\theta}\in \mathbb{R}^{d}$, based on all their training data samples. For simplicity, we assume that users have equal-sized datasets, i.e., $D_i = D$ for all $i\in [N]$.  The typical training goal in FL  can be formally represented by  the following   optimization problem:
\begin{equation}
\label{ma}
\bm {\theta}^* = \arg \min_{\bm {\theta}\in\mathbb{R}^{d}
} \left[ C(\bm {\theta}) := \frac{1}{N} \sum_{i=1}^{N}   C_i(\bm {\theta})\right],
\end{equation}
where  $\bm {\theta}$ is the optimization variable,  $C(\bm {\theta})$ is the global objective function,  $C_i(\bm {\theta})$ is the local loss function of user $i$.    The local loss function of  user $i$ is given by 
\begin{equation}
C_i(\bm {\theta}) = \frac{1}{D} \sum_{(x , y ) \in \mathcal{D}_i } \ell_i(\bm {\theta}, (x , y ) ), 
\end{equation}
where $\ell_i(\bm {\theta}, (x,y))  \in \mathbb{R}$ denotes the  loss function   at a  given data  point $(x_i , y_i )\in \mathcal{D}_i$. The dataset  $\mathcal{D}_i$ at user $i \in [N]$ is sampled from a distribution $\mathcal{P}_i$.   

\begin{figure}
\centering
\includegraphics[width=0.48\textwidth]{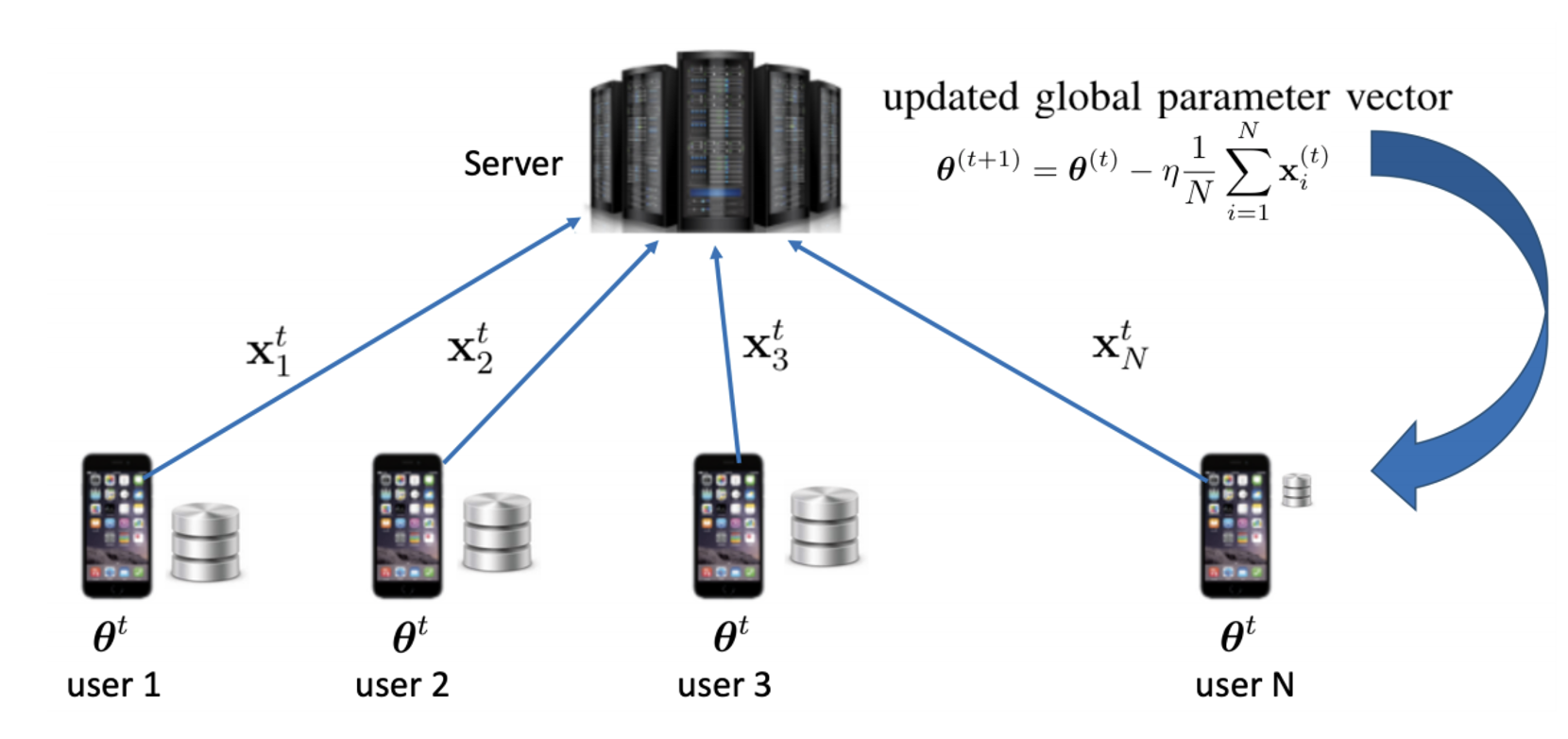}
\caption{The training process in  federated learning.}
\label{SystemModel}
\end{figure}

To solve the optimization problem in \eqref{ma}, an iterative training  procedure is performed  between the server and distributed users, as illustrated in Fig. \ref{SystemModel}. Specifically,
at iteration $t$, the  server firstly sends  the current   global  model  parameters,  $\bm {\theta}^{(t)}$, to  the users.  
User $i\in[N]$ then computes its model update $\mathbf {x}_i^{(t)}$ and sends it to the server. 
After that, the model updates  of the $N$ users are aggregated  by the server to update the global model parameters into $\bm{\theta}^{(t + 1)}$ for the next round according to  
 \begin{equation}
\label{Agg1}
\bm {\theta}^{(t+1)}=\bm {\theta}^{(t)}-\eta^{(t)} \frac{1}{N}   \sum_{i=1}^{N}  \mathbf{x}^{(t)}_i.
\end{equation}
There are two common protocols for computing the model update $\mathbf{x}_i$: \texttt{FedSGD} and \texttt{FedAvg} ~\cite{FedAvg}. Specifically, in \texttt{FedSGD}, each user uses a data batch $\mathcal{B}_i^{(t)}$  of size $B$ sampled uniformly at random from it local dataset $ \mathcal{D}_i$ to compute the model update as follows:

%User   $i\in[N]$ then computes its gradient  (model update) $\mathbf {x}_i^{(t)}$ using a min-batch $\mathcal{B}_i^{(t)}$  of size $B$  sampled uniformly at random from it local dataset $ \mathcal{D}_i$.\footnote{In FL, users can run multiple steps of SGD over their local  data before sending their model updates to the server. In our setting,   we consider the case  where   each user only performs  one step of SGD for simplicity.}  The gradient $\mathbf{x}_i^{(t)}$ at user$i$ is  formally  given as follows
\begin{equation}
\mathbf {x}_i^{(t)} = \frac{1}{B} \sum_{b \in \mathcal{B}_i^{(t)} } g_i( \bm {\theta}^{(t)}, b), 
\end{equation}
where $g_i( \bm {\theta}^{(t)})$  is the stochastic estimate of the gradient $\nabla C_i(\bm {\theta}^{(t)})$ of the local loss function $C_i$ of user $i$ computed based on a random sample $b$ (corresponding to $(x_b,y_b)$)  drawn uniformly   from $\mathcal{D}_i$ without replacement. %\footnote{There are some other works that compute the gradient by sampling the data points uniformly at random with replacements. However, for simplicity we focus on the sampling without  replacment   }
In \texttt{FedAvg}, each user will run $E$ complete  local training rounds over its local dataset $\mathcal{D}_i$ to get its model update $\mathbf {x}_i^{(t)}$. Specifically, during each training round, each user will use all their mini-batches sampled from $\mathcal{D}_i$ to perform multiple stochastic gradient descent steps.

% After getting their model updates, user $i$ ($i\in\mathcal {N}$) sends their model updates  $\mathbf{x}^{(t)}_i$ to the server. The model updates  of the $N$ users are then aggregated  by the server to update the global model $\bm{\theta}^{(t + 1)}$ for the next round according to  
%  \begin{equation}
% \label{Agg1}
% \bm {\theta}^{(t+1)}=\bm {\theta}^{(t)}- \eta^{(t)} \yahya{ \sum_{i=1}^{N} \frac{D_i}{D}} \mathbf{x}^{(t)}_i.
% \end{equation}

\subsection{Secure Aggregation Protocols for Federated Learning} 
% We start by discussing the threat model considered in most of the SA protocols. After that, we formally present the current  privacy guarantee of these protocols. 
Recent works (e.g., ~\cite{NEURIPS2019_60a6c400,geiping2020inverting,yin2021gradients}) have empirically  shown that some of  the local training  data of user $i$   can be reconstructed from the local model update $\mathbf{x}_i$, for $i \in [N]$.  To prevent such data leakage, different 
SA protocols  \cite{aono2017privacy, truex2019hybrid,dong2020eastfly,xu2019hybridalpha,secagg_bell2020secure,secagg_so2021securing,secagg_kadhe2020fastsecagg,zhao2021information,so2021lightsecagg,9712310,mugunthan2019smpai,so2021turbo} have been proposed to  provide  a privacy-preserving FL setting without sacrificing the training performance. In the following, we discuss the threat model used in  these  SA protocols.  
\subsubsection{Threat Model in Secure Aggregation for Federated Learning}\label{sec-threat model}

%Most of SA protocols consider the honest-but-curious model~\cite{cc}. In this threat model,   the server and users  honestly follow the SA protocol as specified. However, they can be curious and try to extract any useful information about the training data of any particular user by analyzing  the    different data  received during the  execution of the protocol. 

Most of SA protocols consider the honest-but-curious model~\cite{cc} with the goal of uncovering users’ data. In this threat model, the server and users honestly follow the SA protocol as specified.  In particular, they  will not modify their model architectures to better suit their attack, nor send malicious model update  that do not represent the actually learned model. However,   the server and the participating users  are assumed to be curious and try to  extract any useful information about the training data of any particular user. The extraction of the information is done by  storing and  analyzing the  different data  received during the  execution of the protocol.

On the other hand,  the threat model in theses SA protocols assumes that the server can collude with any subset of users $\mathcal{T} \subset [N]$ by jointly  sharing  any data that  was used during  the execution of the protocol (including their clear model updates $\mathbf{x}_i$, for all $i \in \mathcal{T}$) that   could help in breaching  the data privacy of any target user $i \in[N]/\mathcal{T}$. Similarly,  this threat model also assumes that users can collude with each other to get information about the training data of other users.  

\subsubsection{Secure Aggregation Guarantees}\label{subsection-SA_guarantee}
 In general,  SA protocols that rely on different encryption techniques;  such as homomorphic encryption \cite{aono2017privacy, truex2019hybrid,dong2020eastfly,xu2019hybridalpha}, and  secure multi-party computing (MPC) \cite{secagg_bell2020secure,secagg_so2021securing,secagg_kadhe2020fastsecagg,zhao2021information,so2021lightsecagg,9712310,mugunthan2019smpai,so2021turbo}, are all  similar in the  encryption procedure in which   each user  encrypts  its own  model update   $\mathbf{y}^{(t)}_i = \text{Enc}(\mathbf{x}^{(t)}_i)$ before sending it to the server. This encryption is done such that these protocols achieve: 1) Correct decoding of the aggregated model under users' dropout; 2) Privacy for the local model update of the users from the encrypted model.  In the following, we formally describe each of these guarantees. \\

\noindent \textbf{Correct decoding.} The encryption  guarantees correct decoding  for the aggregated model of the surviving users even if a subset  $\mathcal{U} \subset[N]$ of the users  dropped out during the protocol execution. In other words, the server should be able to decode 
\begin{equation}
    \text{Dec} \left(\sum_{i \in \mathcal{V}} \mathbf{y}^{(t)}_i \right)= \sum_{i \in \mathcal{V}} \mathbf{x}^{(t)}_i, 
\end{equation}
where $\mathcal{V}$ is the set of surviving users (e.g., $\mathcal{U} \cup \mathcal{V} =[N]$ and  $\mathcal{U} \cap \mathcal{V} = \phi$).

\noindent \textbf{Privacy guarantee.}   Under  the collusion between the server and any strict subset of users $\mathcal{T} \subset [N]$, we have the following 
\begin{equation}\label{eq-SA_guarantee}
   I\left({\{\mathbf{y}}^{(t)}_i\}
   _{i \in[N]}; \{\mathbf{x}^{(t)}_i\}
   _{i \in[N]} \middle  |   \sum_{i=1}^{N} \mathbf{x}^{(t)}_i, \mb{z}_\mathcal{T} \right) = 0,
\end{equation}
where $\mb{z}_\mcal{T}$ is the collection of information at the users in $\mcal{T}$.
In other words,  \eqref{eq-SA_guarantee} guarantees that under a given subset of  colluding users $\mathcal{T}$     with the server,  the  encrypted model updates $\{\mathbf{y}^{(t)}_i\}
   _{i \in[N]}$ leak no information  about the model updates $\{\mathbf{x}^{(t)}_i\}
   _{i \in[N]}$ beyond the aggregated model $\sum_{i=1}^{N} \mathbf{x}^{(t)}_i$.
We note that the upper bound on the size of the colluding set $\mathcal{T}$  such that   \eqref{eq-SA_guarantee} is always guaranteed  has been analyzed in the  different SA protocols. Assuming that  $|\mathcal{T}|\leq \frac{N}{2}$  is widely used in most of the  works   (e.g., \cite{so2021lightsecagg,so2021turbo}). 
\begin{remark}{\rm 
Recently, there have been also some works that enable doing secure model aggregation by using Trusted Execution Environments (TEE) such as  Intel SGX (e.g.,  \cite{9708971,zhang2021shufflefl}).  SGX is a hardware-based security mechanism to protect
applications running on a remote server. These TEE-based works  are also designed to give the same guarantee in \eqref{eq-SA_guarantee}.}
\end{remark}

In the following, we  formally highlight the  weakness of the current  privacy guarantee discussed in \eqref{eq-SA_guarantee}.

\subsubsection{Our Contribution: Guarantees on Privacy Leakage from the  Aggregated Model }

Different  SA protocols guarantee that the server doesn't learn any information about the local model update $\mathbf{x}^{(t)}_i$ of any user  $i$ from the received encrypted updates $\{\mathbf{y}^{(t)}_i\}_{i \in \mcal{N}}$, beyond the aggregated model as formally shown in \eqref{eq-SA_guarantee}. However, it is not clear how much information  the aggregated model update  itself  leaks about  a single user's local dataset $\mathcal{D}_i$. 
 In this work,  we  fill this gap by theoretically  analyzing the following term.  
 \begin{align} \label{eq-aggregate_leake_prelimnies}
  I_{\rm priv/data} = \max_{i \in[N]}\  I\left(\mathcal{D}_i ;  \left \{ \frac{1}{N}  \sum_{i =1}^{N} \mathbf{x}^{(t)}_i  \right\}_{t \in [T]} \right).
\end{align}
The term in \eqref{eq-aggregate_leake_prelimnies} represents how much information the aggregated model over $T$ global  training rounds could leak about the private data $\mathcal{D}_i$ of any user $i \in[N]$. In the following section, we theoretically study this term and discuss how it is impacted by the  different FL system parameters such as model size, number of users , etc. In Section \ref{sec:eval}, we  support our theoretical findings by  empirically evaluating $I_{\rm priv/data}$
in real-world datasets and different neural network architectures.

\section{Theoretical Privacy Guarantees of FL with Secure Aggregation}
\label{sec:privacy_guarantee}

In this section, we  theoretically quantify the privacy leakage in FL when using secure aggregation with the \texttt{FedSGD} protocol.  %then discuss the impact of  FL system parameters on the amount of information leakage.

\subsection{Main Results}\label{sec-Main resuslts}
For clarity, we 
first state our main results under  the  honest-but-curious  threat model discussed in Section \ref{sec-threat model} while assuming that there is  no collusion between the server and  users. We also assume that there is no user  dropout. Later in Section \ref{sec-Impact  of  User Sampling, Users' Dropout, and Collusion}, we discuss the general result with user dropout and  the collusion with the server.

Our central result in this section characterizes the privacy leakage in terms of mutual information for a single round of \texttt{FedSGD}, which for round $t$ is defined as 
\begin{equation}\label{eq:I_priv_round}
I_{\rm priv}^{(t)} = \max_{i\in[N]} I\left(\mathbf{x}^{(t)}_i ;  \sum_{i=1}^N\mb{x}_i^{(t)}\middle|  \left\{\sum_{i=1}^N\mb{x}_i\right\}_{k\in[t-1]} \right)    
\end{equation}
and then extends the privacy leakage bound to multiple rounds. 
Before stating our main result in Theorem~\ref{Thm:main_Thm} below, we first define two key properties of random vectors that will be used in stating  our theorem and  formally state our operational assumptions. 

% \begin{assump}[IID data distribution]  Local dataset $\mathcal{Z}_i$ at node $i$ consists of IID data samples from a distribution $\mathcal{P}_i$, where $\mathcal{P}_i =  \mathcal{P} $ for $ \forall i \in [N]$. This implies that the distribution of the gradients $g_i( \bm {\theta}^{(t)}, b)$, for $i  \in [N]$,  conditioning on the last global model $\bm{\theta}^{(t)}$ are also IID. We denote ${G}^{(t)}_i = G^{(t)}$,  for $i \in [N]$,  to be the conditional   gradient distribution  with  mean $\mu_G^{(t)}$ and covariance matrix $K_G$ at node $i$ in the $t$-th round. 
% \end{assump}

% \begin{defin}[Whitening]\label{def-1}
% A whitening transformation of a random vector  $\mathbf{q}$  with mean $\mu_{q}$ and non-singular covariance matrix $K_{q}$  is a linear transformation that transforms the vector  $\mathbf{q}$  into  of new vector  $\widehat{\mathbf{q}}$ with   the identity covariance matrix. The following linear transformation is a whitening transformation for the vector $\mathbf{q}$
% \begin{equation}\label{eq-7}
%      \widehat{\mathbf{q}} = K_{q}^{-1/2} \left(\mathbf{q} - \mu_{q} \right).
% \end{equation}

% \end{defin}
% of the local model update that different cases for the properties on the gradient vector

\begin{defin}[Independent under whitening]\label{def-2}
We say that a random  vector $\mathbf{v}$ with mean $\mu_{v}$ and non-singular covariance matrix $\mb{K}_{v}$ is \textit{independent under whitening}, if the whitened vector $\widehat{\mb{v}}$ is composed of independent random variables, where     $\widehat{\mathbf{v}} = \mb{K}_{v}^{-1/2} \left(\mathbf{v} - \mu_{v} \right)$.
\end{defin}

\begin{defin}[Uniformly $\sigma$-log concave] \label{def-3}
A random vector $\mathbf{v}$  with covariance $\mb{K}_v$ is {\it uniformly $\sigma$-log concave} if it has a probability density function $e^{-\phi (\mb{v})}$ satisfying $\nabla^2  \phi(\mb{v})  \succeq \mathbf{I}$ and $\exists\ \sigma >0$, such that $\mb{K}_q \succeq \sigma  \mathbf{I}$.  
\end{defin}

\begin{assump}[IID data distribution]  Throughout this section, we consider the case where the local dataset $\mathcal{Z}_i$ are sampled IID from a common distribution, i.e., the local dataset of user $i$ consists of IID data samples from a distribution $\mathcal{P}_i$, where $\mathcal{P}_i =  \mathcal{P}$ for $ \forall i \in [N]$. 
This implies that the distribution of the gradients $g_i( \bm {\theta}^{(t)}, b)$, for $i  \in [N]$,  conditioned on the last global model $\bm{\theta}^{(t)}$ is also IID. For this common conditional distribution, we will denote its mean with $\mu_G^{(t)}$ and the covariance matrix $\mb{K}_{G}^{(t)}$ in the $t$-th round. 
\end{assump}

With the above definitions and using Assumption 1, we can now state our main result  below, which is proved in Appendix~\ref{append:proof_theorem}.

\begin{thm}[Single Round Leakage]\label{Thm:main_Thm}
\textit{
Let $d^* \leq d$ be the rank of the gradient covariance matrix $\mb{K}_{G}^{(t)}$, and let $\mcal{S}_g$ denote the set of subvectors of dimension $d^*$ of $g( \bm {\theta}^{(t-1)}, b)$ that have a non-singular covariance matrices. 
% let $I_{\rm priv}^{(t)}$ be the privacy leakage of a single user's model update $\mb{x}_i^{(t)}$ from the aggregate model $\mb{s}_N^{(t)} = \frac{1}{N}\sum_{i = 1}^N \mathbf{x}^{(t)}_i $ in round $t$, which is defined as
% \[
% I_{\rm priv}^{(t)} = \max_{i\in[N]} I\left(\mathbf{x}^{(t)}_i ;  \mb{s}_N^{(t)}\middle|  \left\{\mb{s}_N^{(k)}\right\}_{k\in[t-1]} \right)
% \]
Under Assumption 1, we can upper bound $I_{\rm priv}^{(t)}$ for \texttt{FedSGD} in the following two cases:\\
\textbf{Case. 1} If  $\exists \bar{g} \in \mcal{S}_g$, such that $\bar{g}$ is independent under whitening (see Def.~\ref{def-2}), and ${\rm E}|\bar{g}_i|^4 < \infty,  \forall i \in [d^*]$, then $\exists\ C_{0,\bar{g}} > 0$, such that
\begin{align}\label{6}
&I_{\rm priv}^{(t)} \leq \frac{C_{0,\bar{g}}\ d^*}{(N-1)B} + \frac{d^*}{2}\log\left(\frac{N}{N-1}\right),
\end{align}
\textbf{Case. 2} If  $\exists \bar{g} \in \mcal{S}_g$, such that $\bar{g}$ is $\sigma$-log concave under whitening (see Def.~\ref{def-3})
then we have that 
\begin{align}\label{eq-case2-thm1}
I_{\rm priv}^{(t)} \leq \frac{d^* C_{1,\bar{g}} - C_{2,\bar{g}}}{(N-1)B\sigma^4}+ \frac{d^*}{2}\log\left(\frac{N}{N-1}\right),
\end{align}
% \end{equation}
 where: the constants $C_{1,\bar{g}} = 2\left( 1+\sigma +\log(2\pi) - \log(\sigma) \right) $ and $C_{2,\bar{g}} = 4\left(h(\bar{g})-\frac{1}{2}\log(|\Sigma_{\bar{g}}|\right)$, with $\Sigma_{\bar{g}}$ being the covariance matrix of the vector $\bar{g}$. 
} 
\end{thm}

\begin{remark}[Simplified bound]\label{remark-simplified-bound}{\rm 
Note that each $\bar{g} \in \mcal{S}_g^{(t)}$ satisfying Case 1 or Case 2 gives an upper bound on $I_{\rm priv}^{(t)}$. Let $\mcal{S}_{g,c}^{(t)}$ be the set of $\bar{g} \in \mcal{S}_g^{(t)}$ satisfying either Case 1 or Case 2.  Then, we can combine these different bounds in Theorem~\ref{Thm:main_Thm} as follows
\begin{align}\label{eq:simplified_round_bounds}
I_{\rm priv}^{(t)} \leq \frac{d^*}{2} \log\left(\!\frac{N}{N{-}1}\!\right) + \dfrac{\displaystyle\min_{\bar{g}\in \mcal{S}_{g,c}^{(t)}} \left\{d^*\widehat{C}_{1,\bar{g}} - \widehat{C}_{2,\bar{g}}\right\} }{(N-1)B},  
\end{align}
where
\begin{align*}
    (\widehat{C}_{1,\bar{g}},\widehat{C}_{2,\bar{g}}) = 
    \begin{cases}
    \left(C_{0,\bar{g}}, 0\right),& \text{if $\bar{g}$ satisfies Case 1}, \\
    \left(\frac{C_{1,\bar{g}}}{\sigma^4}, \frac{C_{2,\bar{g}}}{\sigma^4}\right),& \text{if $\bar{g}$ satisfies Case 2},
    \end{cases}
\end{align*}
where $C_{0,\bar{g}}, C_{1,\bar{g}}$ and $C_{2,\bar{g}}$ are defined as in Theorem~\ref{Thm:main_Thm}.
}
\end{remark}
\begin{remark}{\rm (Why the IID assumption?)
Our main result in Theorem~\ref{Thm:main_Thm} relies on recent results on the  entropic central   \cite{eldan2020clt,bobkov2014berry} for the sum of independent and identically random variables/vectors. Note that the IID assumption in the entropic central limit theorem can be relaxed to independent (but not necessarily identical) distributions, however, in this case, the upper bound will have a complex dependency on the moments of the $N$ distributions in the system. In order to high-light how the privacy guarantee depends on the different system parameters (discussed in the next subsection), we opted to consider the IID setting in our theoretical analysis.
% The reason for providing Theorem 1  under the   assumption of the IID data distribution at the users is that in the proof of this theorem,   we are  adopting some    recent  results in the  entropic central   \cite{eldan2020clt,bobkov2014berry} that are given under the IID sum of random vectors/variables.
}   
\end{remark}

\begin{remark}\yahya{\rm (Independence under whitening)
One of our key assumptions in Theorem~\ref{Thm:main_Thm} is the independence under whitening assumption for stochastic gradient descent (SGD). This assumption is satisfied if the SGD vector can be approximated by a distribution with independent components or by a multivariate Gaussian vector. 
Our adoption of this assumption is motivated by recent theoretical results for analyzing the behaviour of SGD. These results have demonstrated great success in approximating the practical behaviour of SGD, in the context of image classification problems, by modeling the SGD with (i) a non-isotropic Gaussian vector \cite{anisotropic_noise_SGD}, or, (ii) $\alpha$-stable random vectors with independent components \cite{alpha_stable_noise_SGD}. For both these noise models, the independence under whitening assumption in Theorem~\ref{Thm:main_Thm} is valid. However, a key practical limitation for the aforementioned SGD models (and thus of the independence under whitening assumption) is assuming a smooth loss function for learning. This excludes deep neural networks that make use of non-smooth activation and pooling functions (e.g., ReLU and max-pooling). 
}   
\end{remark}

Now using the bounds in Theorem~\ref{Thm:main_Thm}, in the following corollary, we characterize the  privacy leakage of the local training data $\mathcal{D}_i$  of user $i$ after $T$ global  training rounds of \texttt{FedSGD}, which  is defined as 
\begin{equation} \label{eq_short_hand_for_data_leakage_expression}
   I_{\rm priv/data} =  \max_{i\in[N]} I\left(\mcal{D}_i ;  \left \{ \frac{1}{N}  \sum_{i\in [N]} \mathbf{x}^{(t)}_i  \right\}_{t \in [T]} \right),
\end{equation}

\begin{corollary} \label{corollary_data_lekage}
Assuming that users follow the \texttt{FedSGD} training protocol and the same assumptions in Theorem~\ref{Thm:main_Thm}, 
we can derive the upper bound of the privacy leakage $I_{\rm priv/data}$ after $T$ global training rounds of \texttt{FedSGD} in the following two cases:\\
% Given that user $i$ has a  private  training dataset $\mathcal{D}_i$, the privacy leakage of the local data of node $i$ after $T$ global  training round of  \texttt{FedSGD}  can be upper bounded in the following two cases\\
\textbf{Case. 1:} Following the assumptions used in Case 1 in Theorem 1, we get 
\begin{align}
\label{6-a}
 I_{\rm priv/data}\leq T \left[\frac{C_{0,\bar{g}}d^*}{(N-1)B} + \frac{d^*}{2}\log\left(\frac{N}{N-1}\right) \right],
\end{align}
\textbf{Case. 2:} Following the assumptions used in Case 2 in Theorem 1, we get 
\begin{align}
\label{6-b}
I_{\rm priv/data} \leq T \left[\! \frac{d^* C_{1,\bar{g}} - C_{2,\bar{g}}}{(N-1)B\sigma^4}+ \frac{d^*}{2}\log\left(\frac{N}{N{-}1}\right)\! \right].
\end{align}
\end{corollary} 
We prove Corollary~\ref{corollary_data_lekage} in Appendix~\ref{append:proof_Corollary}.
Note that, we can combine the bounds in Corollary~\ref{corollary_data_lekage} similar to the simplification in~\eqref{eq:simplified_round_bounds} from Theorem~\ref{Thm:main_Thm}.

\subsection{Impact of System Parameters}\label{sec-theoratical-system-parameter}
% {\bf Impact of Number of Nodes $N$.}
\subsubsection{Impact of Number of Users (N)}
\label{subsubsec:num_user}
As shown in Theorem \ref{Thm:main_Thm} and Corollary \ref{corollary_data_lekage}, the  upper bounds  on information leakage from the aggregated model update decrease in the number of users $N$. Specifically, the leakage dependency on $N$ is at a rate of $\mathcal{O}(1/N)$.
\subsubsection{Impact of Batch Size (B)}
\label{subsubsec:batch_size}
Theorem 1 and  Corollary \ref{corollary_data_lekage} show that the   information leakage from the aggregated model update  could decrease when increasing the  batch size that is used in  updating the local model of each user. 
\subsubsection{Impact of Model Size (d)}
\label{subsubsec:model_size}
Given our definition of $d^*$ in Theorem 1, where $d^*$ represents the rank of the covariance matrix $K_{G^{(t)}}$ and $d^* \leq d$ ($d$ is the model size),  the leakage given in Theorem \ref{Thm:main_Thm}  and Corollary \ref{corollary_data_lekage} only increases with increasing the  rank of the covariance matrix  of the gradient. This increase happens at a rate of  $\mathcal{O}(d^*)$. In other words, increasing the model size $d$ (especially when the model is overparameterized) does not have a linear impact on the leakage.  The experimental observation in   Section \ref{sec-5} supports these  theoretical findings. 
 \label{remark:model_size}
\subsubsection{Impact of Global Training Rounds (T)}
\label{subsubsec:accum}
Corollary \ref{corollary_data_lekage} demonstrates that the information leakage from the aggregated model update about the private training data of the users increases with  increasing the number of global training rounds. This result reflects the fact as the training proceed, the model at the server start to memorize the training data of the users, and the data of the users is being exposed multiple times by the server as $T$ increases, hence the leakage increases. The increase of the leakage happens at a rate of $\mathcal{O}(T)$. 

\subsection{Impact  of User Dropout,   Collusion, and  User Sampling }\label{sec-Impact  of  User Sampling, Users' Dropout, and Collusion}
In this section, we extend the results given in  Theorem~\ref{Thm:main_Thm} and Corollary~\ref{corollary_data_lekage} to cover  the more   practical FL scenario that consider, user dropout,  the collusion between the server and the users and user sampling. We start by discussing the impact of user dropout and collusion.

\subsubsection{Impact of User Dropout and Collusion with the Server}\label{sec-Impact of users dropout}

Note that, in the case of user dropouts, this is equivalent to a situation where the non-surviving users send a deterministic update of zero. As a result, their contribution can be removed from the aggregated model, and we can, without loss of generality, consider an FL system where only the surviving subset $\mcal{N}_s \subset [N]$ users participate in the system. 

Similarly, when a subset of users colludes with the server, then the server can subtract away their contribution to the aggregated model in order to unmask information about his target user $i$. As a result, we can again study this by considering only the subset of non-colluding (and surviving, if we also consider dropout) users in our analysis. This observation gives us the following derivative of the result in Theorem~\ref{Thm:main_Thm} which can summarized by the following corollary.
% We state the impact of the user dropout and collusion with the server in the following corollary.

\begin{corollary}
In \texttt{FedSGD},  under the assumptions used in Theorem 1, if there is only a subset $\mathcal{N}_s^{(t)} \subset [N]$ of non-colluding and surviving users in the global training  round $t$, then, we have the following bound on $I_{\rm priv}^{(t)}$
\begin{align}\label{eq:simplified_round_bounds_dropout}
I_{\rm priv}^{(t)} \leq \frac{d^*}{2} \log\left(\!\frac{|\mcal{N}_s|}{|\mcal{N}_s|{-}1}\!\right) + \dfrac{\displaystyle\min_{\bar{g}\in \mcal{S}_{g,c}^{(t)}} \left\{d^*\widehat{C}_{1,\bar{g}} - \widehat{C}_{2,\bar{g}}\right\} }{(|\mcal{N}_s|-1)B},  
\end{align}
where the maximization in $I_{\rm priv}^{(t)}$ (given in~\eqref{eq:I_priv_round}) is only over the set of non-colluding surviving and non-colluding users; and the constants $\widehat{C}_{1,\bar{g}}$ and $\widehat{C}_{2,\bar{g}}$ are given in Remark 2. 
\end{corollary}
 
This implies that the per round leakage increases when we have a smaller number of surviving and non-colluding users. Similarly, we can modify the bound in Corollary 1 to take into account user dropout and user collusion by replacing $N$ with $|\mcal{N}_s|$.

\subsubsection{Impact of User Sampling}
In Theorem~\ref{Thm:main_Thm} and Corollary~\ref{corollary_data_lekage}, we assume that all $N$ users in the FL system participate in each training round. If instead $K$ users are chosen each round, then all leakage upper bound will be in terms of $K$, the number of users in each round, instead of $N$. Furthermore, through Corollary~\ref{corollary_data_lekage}, we can develop upper bounds for each user $i$, depending on the number of rounds $T_i$ that the user participated in. For example, taking into account selecting $K$ users in each round denoted by $\mathcal{K}^{(t)}$, then the upper bound in~\eqref{6-a} is modified to give the following information leakage for user $i$
\begin{align}\label{eq:bound_individual_user}
&I_{\rm priv/data}(i) = I\left(\mcal{D}_i ;  \left \{ \frac{1}{K}  \sum_{i\in \mathcal{K}^{(t)}} \mathbf{x}^{(t)}_i  \right\}_{t \in [T]} \right)\nonumber \\
&
\leq T_i \left[\frac{C_{0,\bar{g}}d^*}{(K-1)B} + \frac{d^*}{2}\log\left(\frac{K}{K-1}\right) \right],
\end{align}
where $T_i = K/N$ if the set of $K$ users are chosen independently and uniformly at random in each round.

Thus user sampling would improve the  linear dependence of the leakage on $T$ (Section~\ref{subsubsec:accum}), but increase the per round leakage due to a smaller number of users in each round (Section~\ref{subsubsec:num_user}).

\section{Experimental Setup}\label{sec-5}
\subsection{MI Estimation}

In order to estimate the mutual information in our experiments, we use Mutual Information Neural Estimator (MINE) which is the state-of-the-art method \cite{belghazi2018mine} to estimate the mutual information between two random vectors (see Appendix~\ref{subsec:mine} for more details). In our experiments, at the $t$-th global training round, we use MINE to estimate $I(\mathbf{x}^{(t)}_i;\sum_{i=1}^{N}\mathbf{x}^{(t)}_i | \bm{\theta}^{(t-1)})$, i.e., the mutual information between model update of the $i$-th user $\mathbf{x}^{(t)}_i$ and the aggregated model update from all users $\sum_{i=1}^{N}\mathbf{x}^{(t)}_i$. Our sampling procedure is described as follows: 1) at the beginning of the global training round $t$, each user will first update its local model parameters as the global model parameters $\bm{\theta}^{(t-1)}$. 2) Next, each user shuffles its local dataset. 3) Then, each user will pick a single data batch from its local dataset (if using \texttt{FedSGD}) or use all local data batches (if using \texttt{FedAvg}) to update its local model. 4) Lastly, secure aggregation is used to calculate the aggregated model update. We repeat the above process for $K$ times to get $K$ samples $\{(\mathbf{x}^{(t)}_{i,k};\sum_{i=1}^{N}\mathbf{x}^{(t)}_{i,k})\}_{k=1}^{k=K}$, where $\mathbf{x}^{(t)}_{i,k}$ represents the model update from the $i$-th user in the $k$-th sampling and $\sum_{i=1}^{N}\mathbf{x}^{(t)}_{i,k}$ represents the aggregated model update from the $i$-th user in the $k$-th sampling. Note that we use the $K-th$ (last) sample $\sum_{i=1}^{N}\mathbf{x}^{(t)}_{i,K}$ to update the global model. 
%we simulate this training round for $M$ times to get $M$ samples ($M$=100) for MI estimation. The last sample is used to update global model. We repeat the end-to-end training for 10 times, take the average results, and get the confidence interval with 95\% confidence.

We repeat the end-to-end training and MI estimation multiple times in order to get multiple MI estimates for each training round $t$. We use the estimates for each round to report the average MI estimate and derive the confidence interval (95\%) for the MI estimation\footnote{During our experiments, we observe that the estimated MI does not change significantly across training rounds. Hence, we average the estimated MI across training rounds when reporting our results.}. 
% We choose a confidence interval of 95\% when presenting our results. 

Lastly, when using MINE to estimate MI, we use a fully-connected neural network with two hidden layers each having 100 neurons each as $T_{\theta}$ (see Appendix~\ref{subsec:mine} for more details) and we perform gradient ascent for 1000 iterations to train the MINE network.
%solve optimization problem (\ref{eq:mine}). 
%(\jiang{[Jiang] Maybe add an algorithm block to show the workflow?})

\subsection{Datasets and Models}\label{Models}
\noindent\textbf{Datasets.} We use MNIST and CIFAR10 datasets in our experiments. Specifically, the MNIST dataset contains 60,000 training images and 10,000 testing images, with 10 classes of labels. The CIFAR10 dataset contains 50,000 training images and 10,000 testing images, with 10 classes of labels. For each of the dataset, we randomly split the training data into 50 local datasets with equal size to simulate a total number of 50 users with identical data distribution. Note that we describe how to generate users with non-identical data distribution when we evaluate the impact of user heterogeneity in Section \ref{subsec:hetero}. 
%Moreover, we use Federated EMNIST  (FEMNIST), a benchmark dataset for FL  which is non-IID in its nature.  The FEMNIST consists of  $62$ different classes ($10$ digits, $26$ lowercase, $26$ uppercase) written by different $3500$ writers. 

Moreover, we use MINE to measure the entropy of an individual image in each of these datasets, as an estimate of the maximal potential MI privacy leakage per image. We report that the entropy of an MNIST image is 567 (bits) and the entropy of a CIFAR10 image is 1403 (bits). Note that we will use the entropy of training data to normalize the measured MI privacy leakage in Section \ref{sec:eval}.

\noindent\textbf{Models.} Table \ref{tab:model} reports the models and their number of parameters used in our evaluation. For MNIST dataset, we consider three different models for federated learning. For each of these models, it takes as input a 28$\times$28 image and outputs the probability of 10 image classes. We start by using a simple linear model, with a dimension of 7850. Next, we consider a non-linear model with the same amounts of parameters as the linear model. Specifically, we use a single layer perceptron (SLP), which consists of a linear layer and a ReLU activation function (which is non-linear). Finally, we choose a multiple layer perceptron (MLP) with two hidden layers, each of which contains 100 neurons. In total, it has 89610 parameters. Since the MLP model we use can already achieve more than 95\% testing accuracy on MNIST dataset, we do not consider more complicated model for MNIST.

For the CIFAR10 dataset, we also evaluate three different models for FL. For each of these models, it will take as input an 32$\times$32$\times$3 image and outputs the probability of 10 image classes. Similar to MNIST, the first two models we consider are a linear model and a single layer perceptron (SLP), both of which contains 30720 parameters. The third model we consider is a Convolutional Neural Network (CNN) modified from AlexNet \cite{krizhevsky2012imagenet}, which contains a total of 82554 parameters and is able to achieve a testing accuracy larger than 60\% on CIFAR. We do not consider larger CNN models due to the limited computation resources.
%(since the computation complexity of MINE will increase in order or $O(d)$). 

\begin{table}[!t]
\centering
\begin{tabular}{p{0.6in}<{\centering}|p{0.65in}<{\centering}p{0.65in}<{\centering}p{0.65in}<{\centering}}\hline
\multicolumn{4}{c}{Models for MNIST}\\\hline
Name & Linear & SLP & MLP\\\hline
Size ($d$) & 7890 & 7850 & 89610 \\\hline
\multicolumn{4}{c}{Models for CIFAR10}\\\hline
Name & Linear & SLP & CNN \\\hline
Size ($d$) & 30730 & 30730 & 82554 \\\hline
\end{tabular}
\caption{Models used for MNIST and CIFAR10 datasets. Note that SLP, MLP, and CNN represent Single Layer  Perceptron, Multiple Layer Perceptron, and Convolutional Neural Network, respectively.}
\label{tab:model}
\end{table}
\section{Empirical Evaluation}
\label{sec:eval}
In this section, we empirically evaluate how different FL system parameters affect the MI privacy leakage in SA. Our experiments explore the effect of the system parameters on \texttt{FedSGD}, \texttt{FedAvg} and \texttt{FedProx}~\cite{fedprox_paper}. 
Note that our evaluation results on \texttt{FedSGD} are backed by our theoretical results in Section~\ref{sec:privacy_guarantee}, while our evaluation results on \texttt{FedAvg} and \texttt{FedProx} are purely empirical.

We start by evaluating the impact of the number of users $N$ on the MI privacy leakage for \texttt{FedSGD}, \texttt{FedAvg} and \texttt{FedProx} (see in Section \ref{subsec:num_user}). Then, we evaluate the impact of batch size $B$ on the MI privacy leakage for both \texttt{FedSGD},  \texttt{FedAvg} and \texttt{FedProx} (see in Section \ref{subsec:batch_size}). Next, in Section \ref{subsec:accum}, we measure the accumulative MI privacy leakage across all global training rounds.
\yahya{We evaluate how the local training rounds $E$ for each user will affect the MI privacy leakage for \texttt{FedAvg} and \texttt{FedProx} in Section \ref{subsec:local_epoch}. Finally, the impact of user heterogeneity on the MI privacy leakage for \texttt{FedAvg} is evaluated in Section \ref{subsec:hetero}.} 
% Note that our evaluation results on \texttt{FedSGD} are backed by our theoretical results in Section \ref{sec:privacy_guarantee}, while our evaluation results on \texttt{FedAvg} are purely empirical.

\yahya{We would like to preface by noting that \texttt{FedProx} differs from \texttt{FedAvg} by adding a strongly-convex proximal term to the loss used in \texttt{FedAvg}. Thus, we expect similar dependencies on the number of users $N$, batch-size $B$ and local epochs $E$, when using \texttt{FedAvg} and \texttt{FedProx}.}

% \subsection{Comparison of Privacy Leakage with/without Secure Aggregation}
%%%%%%%%%%%%%%%%%%%%%%%%%%%%%%%%%%
\subsection{Impact of Number of Users (N)}
\label{subsec:num_user}

%%%%%%%%%%%%%%%%%%%%
% Varying N, fedsgd
\begin{figure}[!t]
\begin{subfigure}{.235\textwidth}
    \includegraphics[width=.99\linewidth]{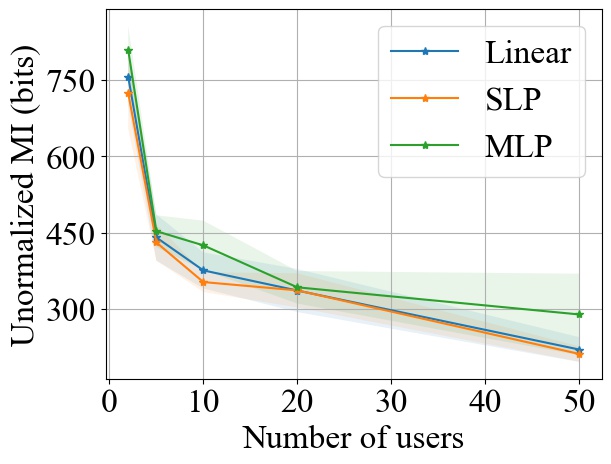}
    \caption{Unnormalized MI, MNIST.}
    \label{fig:fedsgd_n1}
\end{subfigure}
\begin{subfigure}{.235\textwidth}
    \includegraphics[width=.99\linewidth]{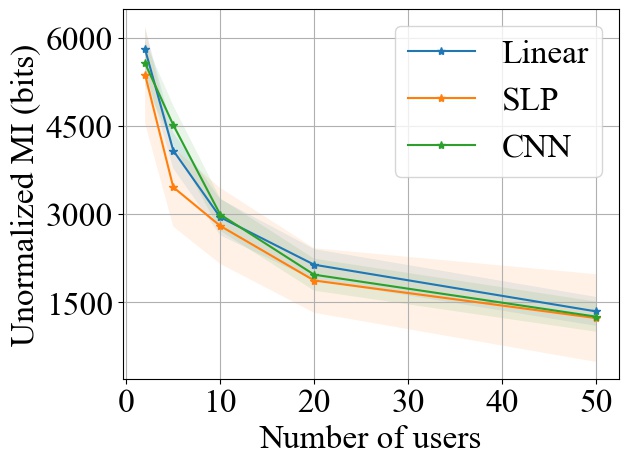}
    \caption{Unnormalized MI, CIFAR10.}
    \label{fig:fedsgd_n2}
\end{subfigure}
\begin{subfigure}{.235\textwidth}
    \includegraphics[width=.99\linewidth]{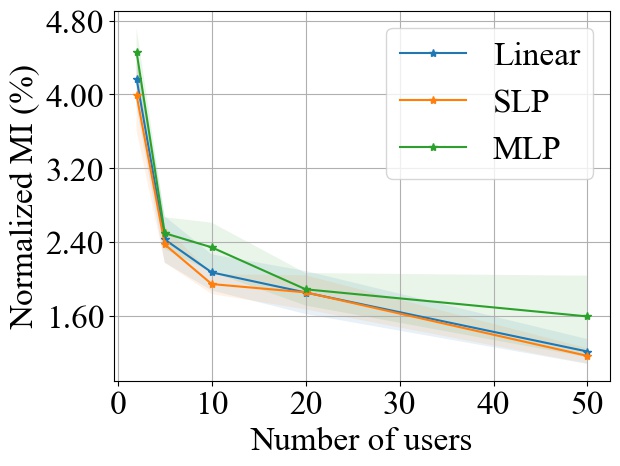}
    \caption{Normalized MI, MNIST.}
    \label{fig:fedsgd_n3}
\end{subfigure}
\begin{subfigure}{.235\textwidth}
    \includegraphics[width=.99\linewidth]{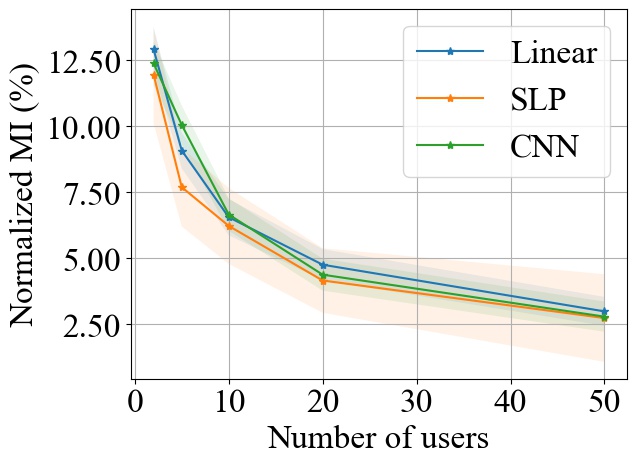}
    \caption{Normalized MI, CIFAR10.}
    \label{fig:fedsgd_n4}
\end{subfigure}
\caption{Impact of the number of users ($N$) when using FedSGD. Note that we set and $B=32$ for all users on both MNIST and CIFAR10 datasets. We normalize the MI by entropy of a single data batch (i.e. $32*567$ for MNIST and $32*1403$ for CIFAR10).} 
\label{fig:num_user_fedsgd}
\vspace{-.1in}
\end{figure}

%%%%%%%%%%%%%%%%%%%%
% Varying N, fedavg
\begin{figure}[!t]
\begin{subfigure}{.235\textwidth}
    \includegraphics[width=.99\linewidth]{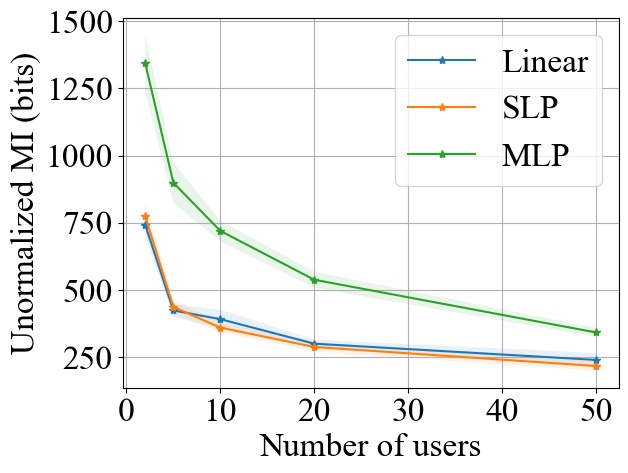}
    \caption{Unnormalized MI, MNIST.}
    \label{fig:fedavg_n1}
\end{subfigure}
\begin{subfigure}{.235\textwidth}
    \includegraphics[width=.99\linewidth]{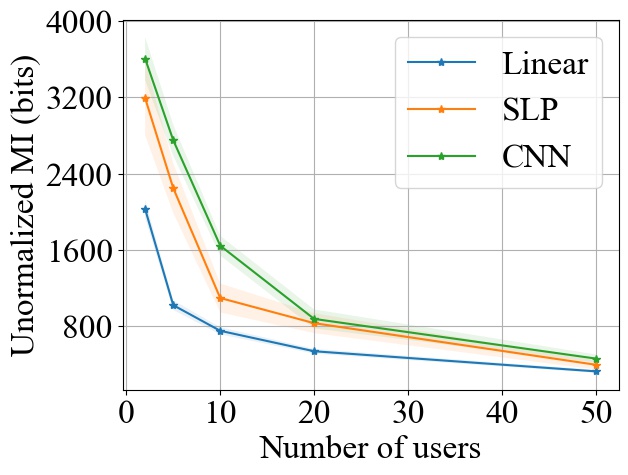}
    \caption{Unnormalized MI, CIFAR10.}
    \label{fig:fedavg_n2}
\end{subfigure}
\begin{subfigure}{.235\textwidth}
    \includegraphics[width=.99\linewidth]{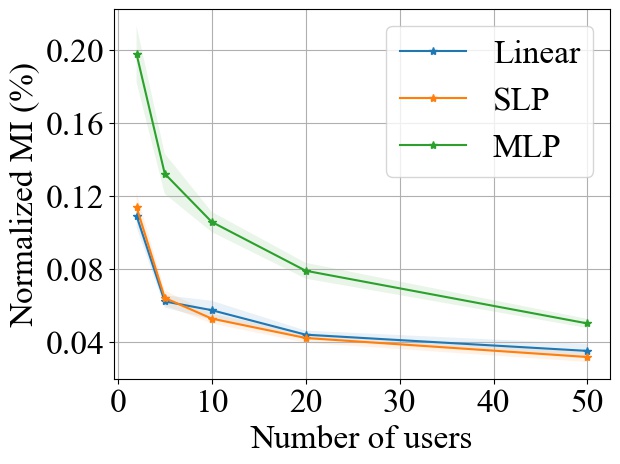}
    \caption{Normalized MI, MNIST.}
    \label{fig:fedavg_n3}
\end{subfigure}
\begin{subfigure}{.235\textwidth}
    \includegraphics[width=.99\linewidth]{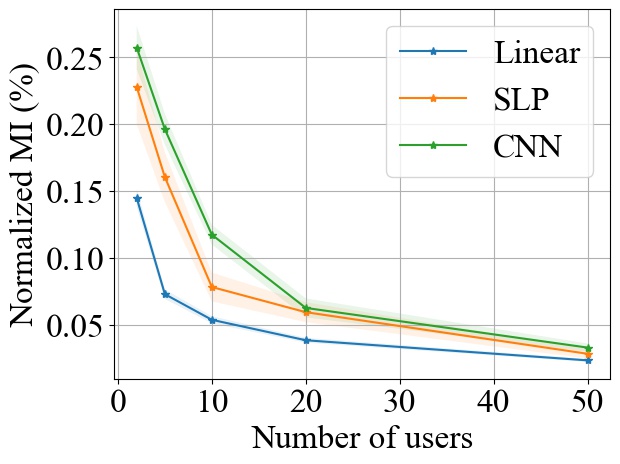}
    \caption{Normalized MI, CIFAR10.}
    \label{fig:fedavg_n4}
\end{subfigure}
\caption{Impact of the number of users ($N$) when using \texttt{FedAvg}. Note that we set $E$=1 and $B=32$ for all users on both MNIST and CIFAR10 datasets. We normalize the MI by entropy of a single data batch (i.e. $1200*567$ for MNIST and $1000*1403$ for CIFAR10).} 
\label{fig:num_user_fedavg}
\vspace{-.1in}
\end{figure}

%%%%%%%%%%%%%%%%%%%%
% Varying N, fedprox
\begin{figure}[!t]
\begin{subfigure}{.235\textwidth}
    \includegraphics[width=.99\linewidth]{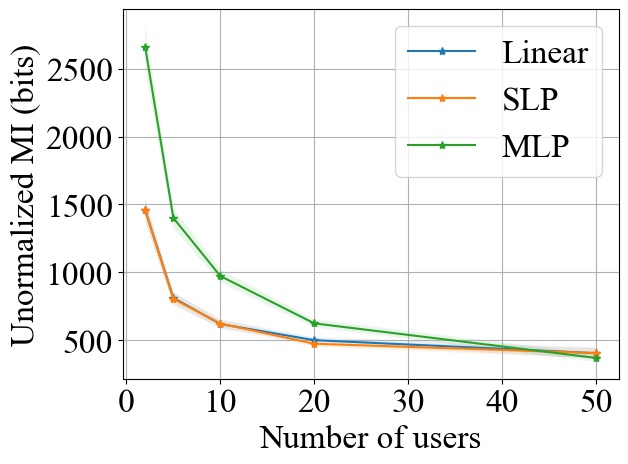}
    \caption{Unnormalized MI, MNIST.}
    \label{fig:fedprox_n1}
\end{subfigure}
\begin{subfigure}{.235\textwidth}
    \includegraphics[width=.99\linewidth]{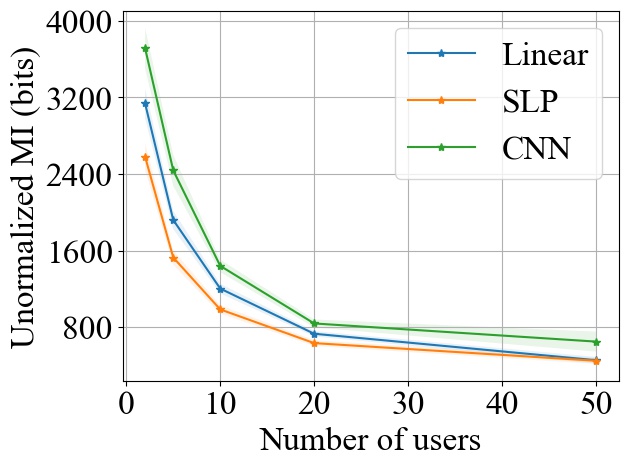}
    \caption{Unnormalized MI, CIFAR10.}
    \label{fig:fedprox_n2}
\end{subfigure}
\begin{subfigure}{.235\textwidth}
    \includegraphics[width=.99\linewidth]{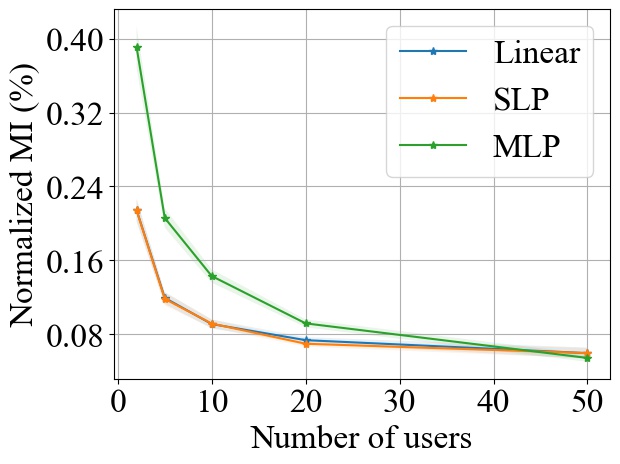}
    \caption{Normalized MI, MNIST.}
    \label{fig:fedprox_n3}
\end{subfigure}
\begin{subfigure}{.235\textwidth}
    \includegraphics[width=.99\linewidth]{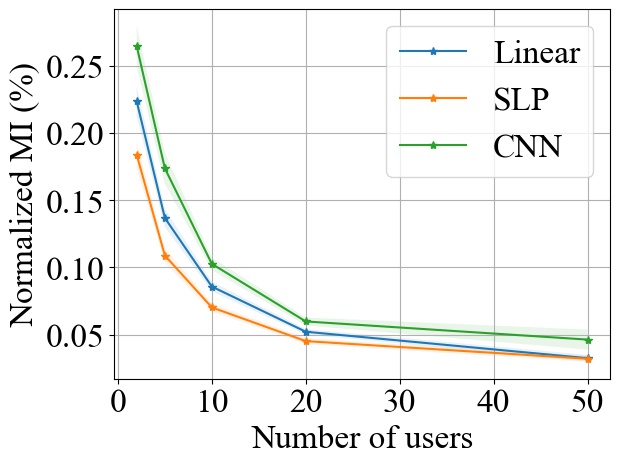}
    \caption{Normalized MI, CIFAR10.}
    \label{fig:fedprox_n4}
\end{subfigure}
\caption{Impact of the number of users ($N$) when using \texttt{FedProx}. Note that we set $E$=1 and $B=32$ for all users on both MNIST and CIFAR10 datasets. We normalize the MI by entropy of a single data batch (i.e. $1200*567$ for MNIST and $1000*1403$ for CIFAR10).} 
\label{fig:num_user_fedprox}
\vspace{-.1in}
\end{figure}

\noindent\textbf{FedSGD.} Fig. \ref{fig:num_user_fedsgd} shows the impact of varying $N$ on MI privacy leakage in \texttt{FedSGD}, where the number of users is chosen from $\{2,5,10,20,50\}$, and we measure the MI privacy leakage of different models on both MNIST and CIFAR10 datasets. We observe that increasing the number of users participating in FL using \texttt{FedSGD} will decrease the MI privacy leakage in each global training round (see Fig. \ref{fig:fedsgd_n1} and \ref{fig:fedsgd_n2}), which is consistent with our theoretical analysis in Section \ref{subsubsec:num_user}. Notably, as demonstrated in Fig. \ref{fig:fedsgd_n3} and \ref{fig:fedsgd_n4}, the percentile of MI privacy leakage (i.e. normalized by the entropy of a data batch) can drop below 2\% for MNIST and 5\% for CIFAR10 when there are more than 20 users. 
%Furthermore, we observe that increasing model size $d$ will increase the MI leakage during each global training round. However, the increase rate of MI leakage is smaller than the increase rate of $d$. This is expected since the upper bound of MI privacy leakage is proportional to $d^*$ (i.e. the rank of the covariance of matrix as proved in Theorem \ref{Thm:main_Thm}),  which will not increase linearly with $d$ especially for overparameterized neural networks (see Section \ref{subsubsec:model_size}). Finally, we observe that the MI privacy leakage on CIFAR10 is generally higher than that on MNIST. Since the input images on CIFAR10 have higher dimension than the images on MNIST, larger model size are required during training. Therefore, we expect that the MI privacy leakage on CIFAR10 is higher than that on MNIST.

\noindent\textbf{FedAvg.} Fig. \ref{fig:num_user_fedavg} shows the impact of varying $N$ on MI privacy leakage in \texttt{FedAvg}. Similar to the results in \texttt{FedSGD}, as the number of users participating in \texttt{FedAvg} increases, the MI privacy leakage in each global training round will decrease (see Fig. \ref{fig:fedavg_n1} and \ref{fig:fedavg_n2}), and the decreasing rate is approximately $\mathcal{O}(N)$. Moreover, as shown in Fig. \ref{fig:fedavg_n3} and \ref{fig:fedavg_n4}, the percentile of MI privacy leakage drops below 0.1\% on both MNIST and CIFAR10 when there are more than 20 users participating in FL. It is worth noting that we normalize the MI by the entropy of the whole training dataset in \texttt{FedAvg} instead of the entropy of a single batch, since users will iterate over all their data batches to calculate their local model updates in \texttt{FedAvg}. Therefore, although we observe that the unnormalized MI is comparable for \texttt{FedSGD} and \texttt{FedAvg}, the percentile of MI privacy leakage in \texttt{FedAvg} is significantly smaller than that in \texttt{FedSGD}. 

\noindent\textbf{FedProx.} \yahya{Similar to \texttt{FedAvg}, Fig.~\ref{fig:num_user_fedprox} shows how the MI privacy leakage with \texttt{FedProx} varies with the number of users $N$. As the number of users increase, the MI privacy leakage decreases in each training round at an approximate rate of $O(N)$. With more than 20 participating users, the percentile of MI leakage drops below 0.12\% under both MNIST and CIFAR10. Same as \texttt{FedAvg}, we normalize the MI privacy leakage by the entropy of the whole training dataset of a single user.
% ~
% Note that \texttt{FedProx} represents an application of \texttt{FedAvg} on the  original loss function in addition to the strongly convex proximal term. This explains why \texttt{FedProx} and \texttt{FedAvg} exhibit similar trends and closely similar MI leakage percentiles.
}

In conclusion, while our theoretical analysis on the impact of $N$ in Section \ref{subsubsec:num_user} is based on the assumption that the \texttt{FedSGD} protocol is used, our empirical study shows that it \yahya{holds not only in \texttt{FedSGD} but also in  \texttt{FedAvg} and \texttt{FedProx}}.

\subsection{Impact of Model Size (d)}
\label{subsec:model_size}
\noindent\textbf{FedSGD.} From Fig. \ref{fig:num_user_fedsgd}, 
%shows the impact of varying $N$ on MI privacy leakage in \texttt{FedSGD}, where the number of users is chosen from $\{2,5,10,20,50\}$, and we measure the MI privacy leakage of different models on both MNIST and CIFAR10 datasets. 
%We observe that increasing the number of users participating in FL using \texttt{FedSGD} will decrease the MI privacy leakage in each global training round (see Fig. \ref{fig:fedsgd_n1} and \ref{fig:fedsgd_n2}), which is consistent with our theoretical analysis in Section \ref{subsubsec:num_user}. Notably, as demonstrated in Fig. \ref{fig:fedsgd_n3} and \ref{fig:fedsgd_n4}, the percentile of MI privacy leakage (i.e. normalized by the entropy of a data batch) can drop below 2\% for MNIST and 5\% for CIFAR10 when there are more than 20 users. Furthermore, 
we observe that increasing model size $d$ will increase the MI leakage during each global training round. However, the increase rate of MI leakage is smaller than the increase rate of $d$. This is expected since the upper bound of MI privacy leakage is proportional to $d^*$ (i.e. the rank of the covariance of matrix as proved in Theorem \ref{Thm:main_Thm}),  which will not increase linearly with $d$ especially for overparameterized neural networks (see Section \ref{subsubsec:model_size}). Finally, we observe that the MI privacy leakage on CIFAR10 is generally higher than that on MNIST. Since the input images on CIFAR10 have higher dimension than the images on MNIST, larger model size are required during training. Therefore, we expect that the MI privacy leakage on CIFAR10 is higher than that on MNIST.

\noindent\textbf{FedAvg and FedProx.} \yahya{As shown in Fig. \ref{fig:num_user_fedavg} and Fig. \ref{fig:num_user_fedprox}, 
%shows the impact of varying $N$ on MI privacy leakage in \texttt{FedAvg}. Similar to the results in \texttt{FedSGD}, as the number of users participating in \texttt{FedAvg} increases, the MI privacy leakage in each global training round will decrease (see Fig. \ref{fig:fedavg_n1} and \ref{fig:fedavg_n2}), and the decreasing rate is approximately $\mathcal{O}(N)$. Moreover, as shown in Fig. \ref{fig:fedavg_n3} and \ref{fig:fedavg_n4}, the percentile of MI privacy leakage drops below 0.1\% on both MNIST and CIFAR10 when there are more than 20 users participating in FL. It is worth noting that we normalize the MI by the entropy of the whole training dataset in \texttt{FedAvg} instead of the entropy of a single batch, since users will iterate over all their data batches to calculate their local model updates in \texttt{FedAvg}. Therefore, although we observe that the unnormalized MI is comparable for \texttt{FedSGD} and \texttt{FedAvg}, the percentile of MI privacy leakage in \texttt{FedAvg} is significantly smaller than that in \texttt{FedSGD}. Lastly, 
increasing the model size will also have a sub-linear impact on the increase of  the MI privacy leakage in \texttt{FedAvg} and \texttt{FedProx}, which is consistent with our results in \texttt{FedSGD}.}

% The above findings suggest that large number of user can help to reduce the MI privacy leakage, and we should use FedAvg.
%In conclusion, while our theoretical analysis on the impact of $d$ in Section \ref{subsubsec:num_user} and Section \ref{sec:privacy_guarantee} is based on the assumption that the \texttt{FedSGD} protocol is used, our empirical study shows that it can hold in both \texttt{FedSGD} and \texttt{FedAvg}.
%%%%%%%%%%%%%%%%%%%%%%%%%%%%%%%%%%
\subsection{Impact of Batch Size (B)}
\label{subsec:batch_size}
%%%%%%%%%%%%%%%%%%%%
% Varying B, fedsgd
\begin{figure}[!t]
% \begin{subfigure}{.24\textwidth}
%     \includegraphics[width=.99\linewidth]{figs/impact_B/results_cmp_bs_bs_high_avg_sgd.jpg}
%     \caption{Unnormalized MI on MNIST.}
%     \label{fig:fedsgd_b1}
% \end{subfigure}
% \begin{subfigure}{.24\textwidth}
%     \includegraphics[width=.99\linewidth]{figs/impact_B/results_cmp_bs_avg_bs_high_avg_sgd_cifar10.jpg}
%     \caption{Unnormalized MI on CIFAR10.}
%     \label{fig:fedsgd_b2}
% \end{subfigure}
\begin{subfigure}{.235\textwidth}
    \includegraphics[width=.99\linewidth]{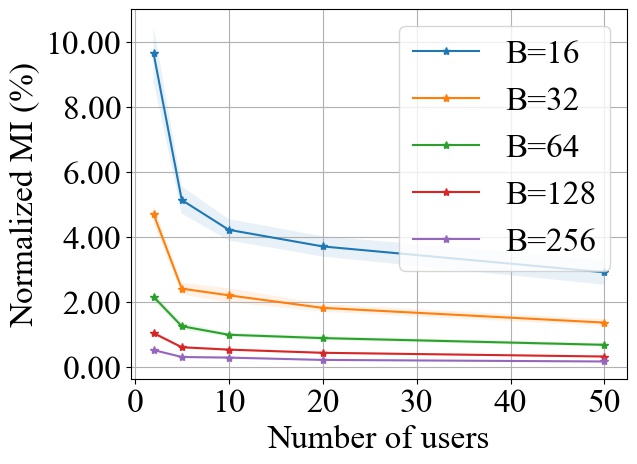}
    \caption{Normalized MI, MNIST.}
    \label{fig:fedsgd_b3}
\end{subfigure}
\begin{subfigure}{.235\textwidth}
    \includegraphics[width=.99\linewidth]{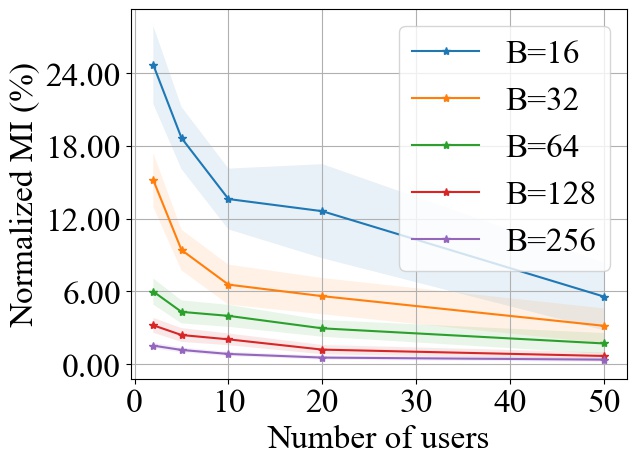}
    \caption{Normalized MI, CIFAR10.}
    \label{fig:fedsgd_b4}
\end{subfigure}
\caption{Impact of batch size ($B$) when using FedSGD. The MI is normalized by the entropy of a data batch, which is proportional to the batch size $B$ (i.e. $B*567$ for MNIST and $B*1403$ for CIFAR10).} 
\label{fig:bs_fedsgd}
\vspace{-.1in}
\end{figure}
%%%%%%%%%%%%%%%%%%%%
% Varying B, fedavg
\begin{figure}[!t]
\begin{subfigure}{.235\textwidth}
    \includegraphics[width=.99\linewidth]{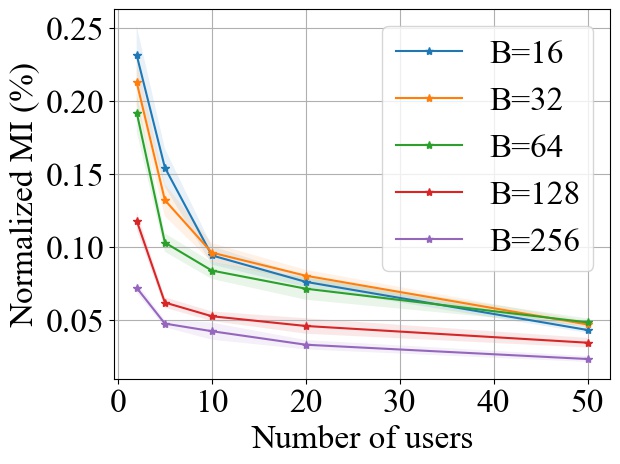}
    \caption{Normalized MI, MNIST.}
    \label{fig:fedavg_b1}
\end{subfigure}
\begin{subfigure}{.235\textwidth}
    \includegraphics[width=.99\linewidth]{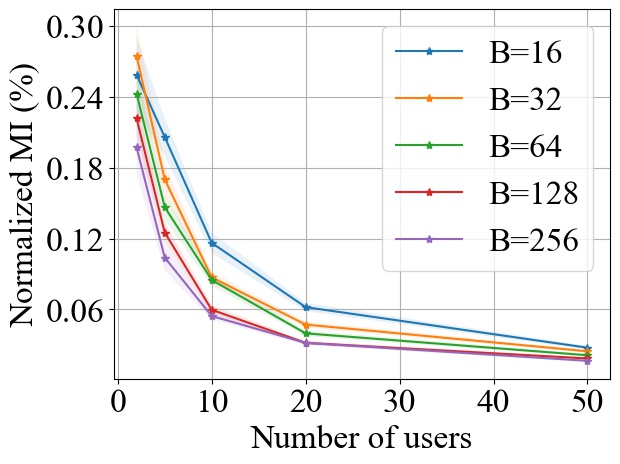}
    \caption{Normalized MI, CIFAR10.}
    \label{fig:fedavg_b2}
\end{subfigure}
\caption{Impact of batch size ($B$) when using FedAvg. The MI is normalized by the entropy of a user's local dataset, which is a constant (i.e. $1200*567$ for MNIST and $1000*1403$ for CIFAR10).} 
\label{fig:bs_fedavg}
\vspace{-.1in}
\end{figure}

%%%%%%%%%%%%%%%%%%%%
% Varying B, fedprox
\begin{figure}[!t]
\begin{subfigure}{.235\textwidth}
    \includegraphics[width=.99\linewidth]{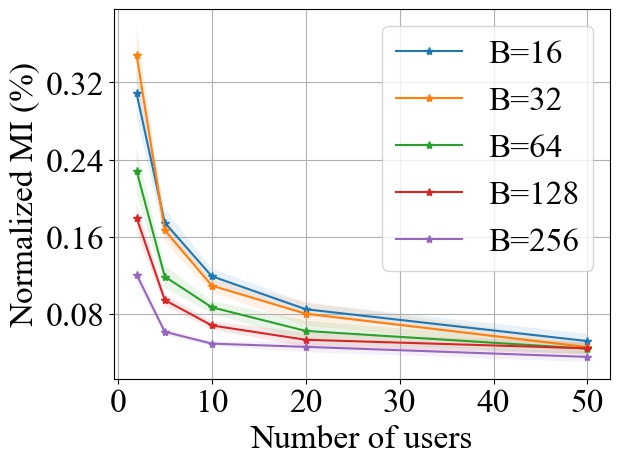}
    \caption{Normalized MI, MNIST.}
    \label{fig:fedprox_b1}
\end{subfigure}
\begin{subfigure}{.235\textwidth}
    \includegraphics[width=.99\linewidth]{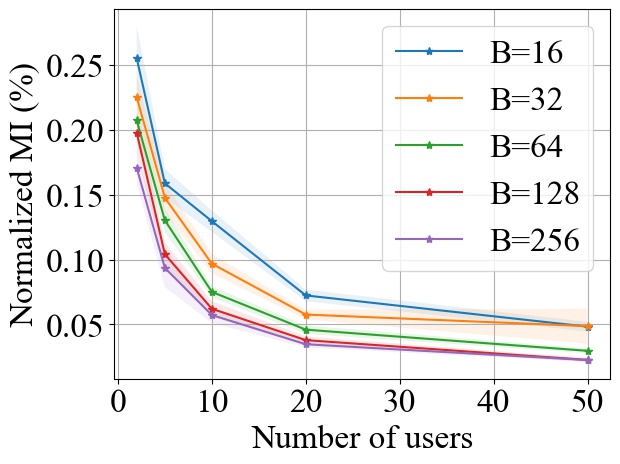}
    \caption{Normalized MI, CIFAR10.}
    \label{fig:fedprox_b2}
\end{subfigure}
\caption{Impact of batch size ($B$) when using FedProx. The MI is normalized by the entropy of a user's local dataset, which is a constant (i.e. $1200*567$ for MNIST and $1000*1403$ for CIFAR10).} 
\label{fig:bs_fedprox}
\vspace{-.1in}
\end{figure}

\noindent\textbf{FedSGD.} Fig. \ref{fig:bs_fedsgd} shows the impact of varying $B$ on the normalized MI privacy leakage in \texttt{FedSGD}, where the batch size is chosen from  $\{16,32,64,128,256\}$ and we use MLP model on MNIST and CNN model on CIFAR10 during experiments. Note that we normalize the MI by the entropy of a single data batch used in each training round, which is proportional to the batch size $B$. On both MNIST and CIFAR10 datasets, we consistently observe that increasing $B$ will decrease the MI privacy leakage in \texttt{FedSGD}, and the decay rate of MI is inversely proportional to batch size $B$. As demonstrated in Fig. \ref{fig:bs_fedsgd}, when there are more than 20 users, the percentile of MI privacy leakage for a single training round can be around 4\% on MNIST and 12\% on CIFAR10 with batch size 16. However, such leakage can drop to less 1\% on both MNIST and CIFAR10 with batch size 256, which is significantly reduced. 

\noindent\textbf{FedAvg and FedProx.} \yahya{Fig. \ref{fig:bs_fedavg} and Fig.~\ref{fig:bs_fedprox} show the impact of varying the batch size $B$ on MI privacy leakage in \texttt{FedAvg} and \texttt{FedProx}, respectively, following the same experimental setup as in Fig. \ref{fig:bs_fedsgd}. Since in both \texttt{FedAvg} and \texttt{FedProx}, each user will transverse their whole local dataset in each local training round, we normalize the MI by the entropy of the target user's local training dataset. As shown in Fig. \ref{fig:bs_fedavg} and Fig. \ref{fig:bs_fedprox}, the impact of $B$ in \texttt{FedAvg} and \texttt{FedProx} is relatively smaller than that in \texttt{FedSGD}. However, we can still observe that increasing $B$ can decrease the MI privacy leakage in both \texttt{FedAvg} and \texttt{FedProx}. For example, with 20 users participating in \texttt{FedAvg}, the percentile of MI privacy leakage at each training round can drop from 0.8\% to 0.3\% when the batch size increases from 16 to 256, achieving a reduction in privacy leakage by  a factor  of more than 2$\times$.
Similarly, in \texttt{FedProx}, this causes a decrease in the MI privacy leakage from 0.09\% to 0.04\% when the batch size increases from 16 to 256.}

% The above findings suggest that we can use large batch size in FL systems to mitigate the MI privacy leakage of individual users, and we should use FesAvg.
In conclusion, we observe that increasing the batch size $B$ can decrease the MI privacy leakage from the aggregated model update in \yahya{\texttt{FedSGD}, \texttt{FedAvg} and \texttt{FedProx}} which verifies our theoretical analysis in Section \ref{subsubsec:model_size}.  

\subsection{Accumulative MI leakage}
\label{subsec:accum}
To evaluate how the accumulative MI privacy leakage will accumulate with the number of training round $T$, we measure the MI between training data and the aggregated model updates across training round. Specifically, given a local training dataset sample $\mcal{D}_i$, we will concatenate the aggregated model updates $\{\frac{1}{N}\sum_{i\in\mathcal{N}}\mathbf{x}_i^{(t)}\}_{t\in[T]}$ across $T$ training rounds in a single vector with dimension $d*T$. By randomly generating $\mathcal{D}_i$ for the target user for $K$ times, we can get $K$ concatenated aggregated model update vectors. Then, we use MINE to estimate $I(\mathcal{D}_i;\{\frac{1}{N}\sum_{i\in\mathcal{N}}\mathbf{x}_i^{(t)}\}_{t\in[T]})$ with these $K$ dataset and concatenated model update samples. 

\begin{figure}[!t]
\centering
\begin{subfigure}{.235\textwidth}
    \includegraphics[width=.99\linewidth]{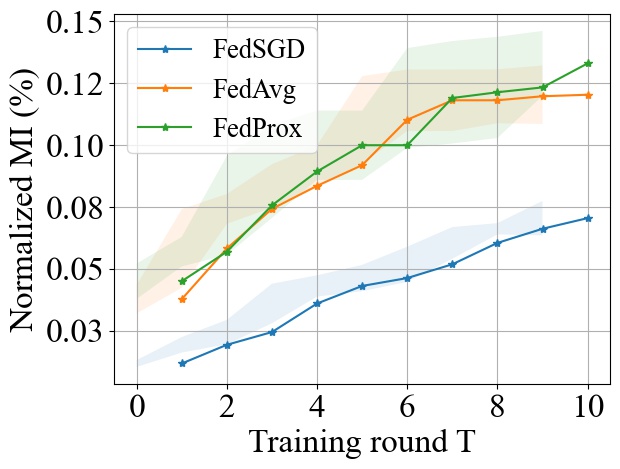}
    \caption{Normalized accumulative MI, MNIST.}
    \label{fig:fedsgd_accum}
\end{subfigure}
\begin{subfigure}{.235\textwidth}
    \includegraphics[width=.99\linewidth]{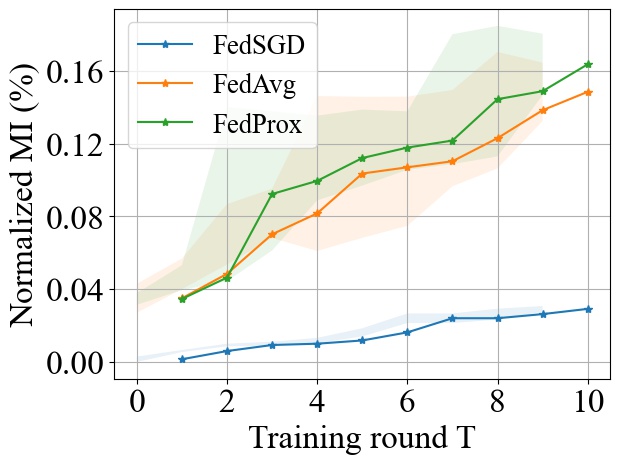}
    \caption{Normalized accumulative MI, CIFAR10.}
    \label{fig:fedavg_accum}
\end{subfigure}
\caption{Accumulative MI privacy leakage on MNIST and CIFAR10 datasets. Note that we normalize the MI by the entropy of each user's local dataset, which will not change with $T$. We use the linear model for both MNIST and CIFAR10 datasets.} 
\label{fig:accum}
\vspace{-.1in}
\end{figure}

As illustrated in Fig. \ref{fig:accum}, the MI privacy leakage will accumulate linearly as we increase the global training round $T$ on both MNIST and CIFAR dataset, which is consistent with our theoretical results in Section \ref{subsubsec:accum}. That also says, by reducing the times of local model aggregation, the MI privacy leakage of secure aggregation will be reduced. In practice, we can consider using client sampling to reduce the participation times of each client in FL, such that the accumulative MI leakage of individual users can be reduced. Moreover, we can also consider increasing the number of local averaging as much as possible to reduce the aggregation times for local model updates.
% Moreover, we can also consider increasing the number of local epochs as much as possible to reduce the number of aggregations for local model updates.

\begin{figure}[!t]
\centering
\begin{subfigure}{.235\textwidth}
    \includegraphics[width=.99\linewidth]{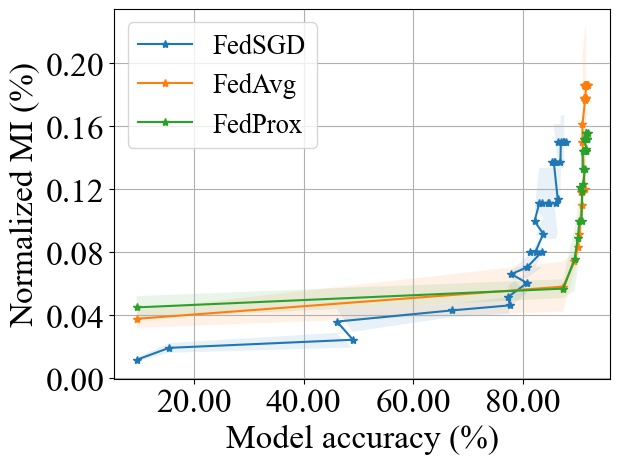}
    \caption{MNIST}
    \label{fig:fedsgd_accum_acc}
\end{subfigure}
\begin{subfigure}{.235\textwidth}
    \includegraphics[width=.99\linewidth]{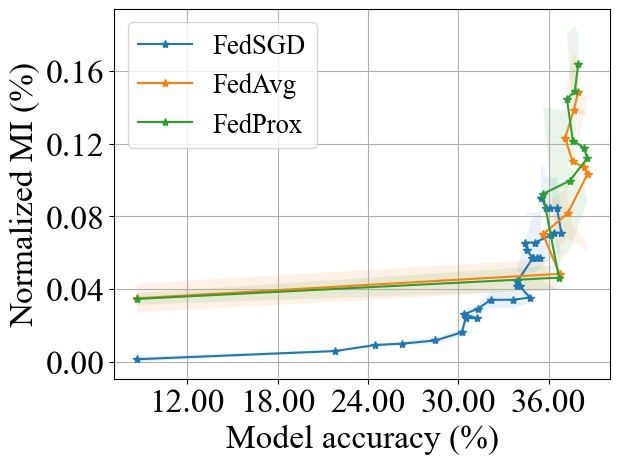}
    \caption{CIFAR}
    \label{fig:fedavg_accum_acc}
\end{subfigure}
\caption{Accumulative MI privacy leakage vs model accuracy of different FL algorithms. Note that we use a linear model for case study and normalize the MI by the entropy of each user's local dataset.} 
\label{fig:accum_acc}
\vspace{-.1in}
\end{figure}

% \yahya{From Fig.~\ref{fig:accum}, we can see that FedSGD, FedAvg and FedProx lead to comparable accumulative MI privacy leakage for a given number of rounds.} \kostas{COMMENT: I do NOT see this in the plots. Likely there is a typo in 9(c) with the two datasets being interchanged?} \yahya{However, these different aggregation algorithms can result in different convergence speeds to a target accuracy (FedSGD in particular has the slowest convergence). To highlight the effect of convergence rate on the accumulative MI privacy leakage, we show the amount of MI leakage incurred for reaching a target model accuracy in Table~\ref{tab:converge}.}
\yahya{Although, the three aggregation algorithms exhibit a similar trend with $T$, these algorithms can result in different convergence speeds to a target accuracy.
To highlight the effect of convergence rate on the accumulative MI privacy leakage, we show, in Fig.~\ref{fig:accum_acc}, how the accuracy changes with the amount of MI leakage incurred for the three algorithms during the training process up to a maximum of 30 training rounds for FedSGD. We observe that although FedSGD achieves lower MI leakage for a fixed number of rounds (see Fig.~\ref{fig:accum}), its slow convergence rate will make it suffer from more leakage before reaching a target accuracy rate. For example, given a target accuracy of 85\% on the MNIST dataset, both FedAvg and FedProx achieve the target accuracy with 0.058\% and 0.057\% leakage while FedSGD will reach 85\% accuracy in later rounds resulting in an accumulative MI leakage of 0.11\% (even with smaller leakage per round).}
%
%\kostas{COMMENT: what is the rational of selecting 89\% and 37\% accuracy for these two datasets? Why not show a plot with the leakage as a function of the accuracy for an accuracy varying say from 50 to 90 in steps of 10, 3 lines per plot for each FedXXX version?}
% \yahya{We notice that FedProx and FedAvg need 5 training rounds ($T$) to reach the target accuracy and as a consequence, achieve lower accumulative MI privacy leakage compared to FedSGD
% %(FedProx is the best for CIFAR10 and FedAvg is the best for MNIST)
% . In comparison, FedSGD requires 50 training rounds to achieve the target accuracy, resulting in a higher accumulative privacy leakage on both MNIST and CIFAR10.}

%%%%%%%%%%%%%%%%%%%%
% Varying training round

\subsection{Impact of Local Training Epochs (E)}
\label{subsec:local_epoch}
%%%%%%%%%%%%%%%%%%%%
% Varying E, fedavg
\begin{figure}[!t]
\begin{subfigure}{.235\textwidth}
    \includegraphics[width=.99\linewidth]{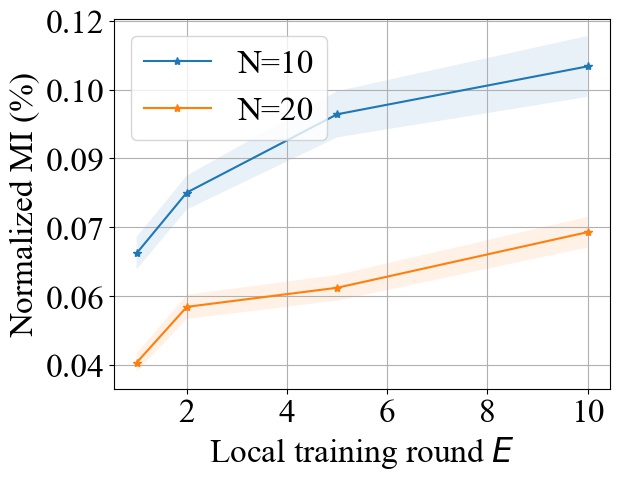}
    \caption{Normalized MI, MNIST.}
    \label{fig:fedavg_e1}
\end{subfigure}
\begin{subfigure}{.235\textwidth}
    \includegraphics[width=.99\linewidth]{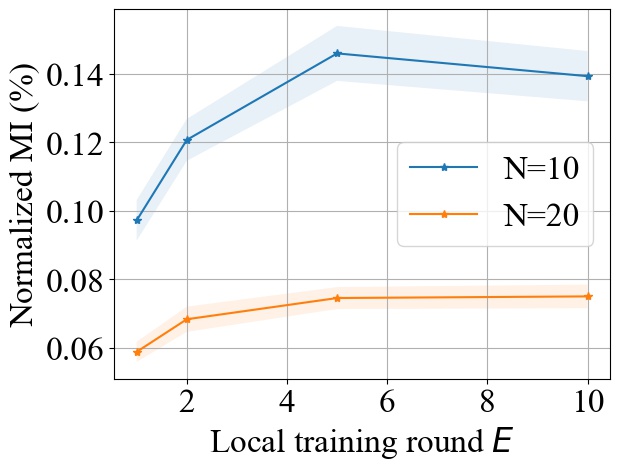}
    \caption{Normalized MI, CIFAR10.}
    \label{fig:fedavg_e2}
\end{subfigure}
\caption{Impact of the local training round ($E$) when using FedAvg. We normalize the MI by the entropy of each user's local dataset, and we consider $N\in\{10,20\}$.} 
\label{fig:ep_fedavg}
\vspace{-.1in}
\end{figure}
% Varying E, FedProx
\begin{figure}[!t]
\begin{subfigure}{.235\textwidth}
    \includegraphics[width=.99\linewidth]{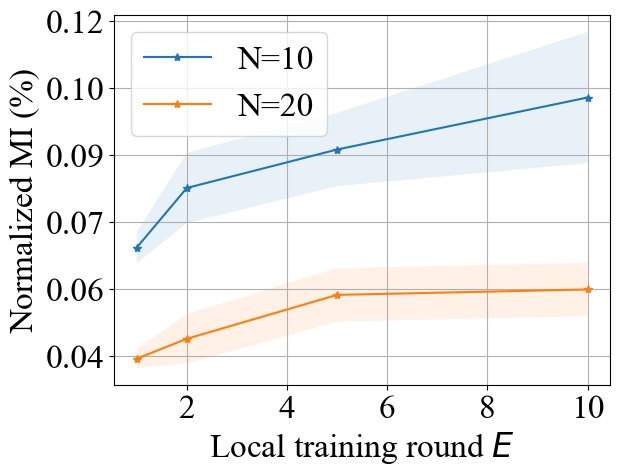}
    \caption{Normalized MI, MNIST.}
    \label{fig:fedprox_e1}
\end{subfigure}
\begin{subfigure}{.235\textwidth}
    \includegraphics[width=.99\linewidth]{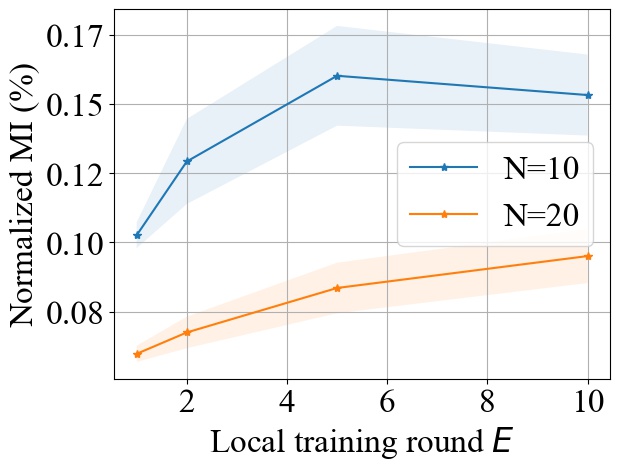}
    \caption{Normalized MI, CIFAR10.}
    \label{fig:fedprox_e2}
\end{subfigure}
\caption{Impact of the local training round ($E$) when using FedProx. We normalize the MI by the entropy of each user's local dataset, and we consider $N\in\{10,20\}$.} 
\label{fig:ep_fedprox}
\vspace{-.1in}
\end{figure}
%%%%%%%%%%%%%%%%%%%%
Fig. \ref{fig:ep_fedavg} shows the impact of varying the number of local training epochs $E$ on MI privacy leakage in \texttt{FedAvg} on both MNIST and CIFAR10 datasets. We select $E$ from $\{1,2,5,10\}$ and $N$ from $\{10,20\}$, and we consider MLP model for MNIST and CNN model for CIFAR10.  We observe that increasing the local training round $E$ will increase the MI privacy leakage in \texttt{FedAvg}. An intuitive explanation is that with more local epochs, the local model updates become more biased towards the user's local dataset, hence it will potentially leak more private information about users' and make it easier for the server to infer the individual model update from the aggregated update. However, as shown in Fig. \ref{fig:ep_fedavg}, increasing the local epochs $E$ will not have a linear impact on the increase of MI privacy leakage. As $E$ increases, the increase rate of MI privacy leakage becomes smaller. 
%However, as the number of users increase, such effects can be mitigated, since aggregating model updates from more user will offer more protection for individual model update from each user, and hence reduce the potentially increased MI privacy leakage caused by more local averaging.

\yahya{Similar to \texttt{FedAvg}, we observe from Fig.~\ref{fig:ep_fedprox} that the local training epochs $E$ has a sub-linear impact on the MI privacy leakage when using \texttt{FedProx}. As aforementioned, this can be attributed to the fact that \texttt{FedProx} represents an application of \texttt{FedAvg} with the original loss function in addition to a convex regularization term.}
% (See Appendix~\ref{subsec:FedProx})}. 
% \kostas{COMMENT: did you meant to say "...with the addition of a convex proximal term." ?}

\subsection{Impact of Data Heterogeneity}
\label{subsec:hetero}
%%%%%%%%%%%%%%%%%%%%
% Varying E, fedavg
\begin{figure}[!t]
\begin{subfigure}{.235\textwidth}
    \includegraphics[width=.99\linewidth]{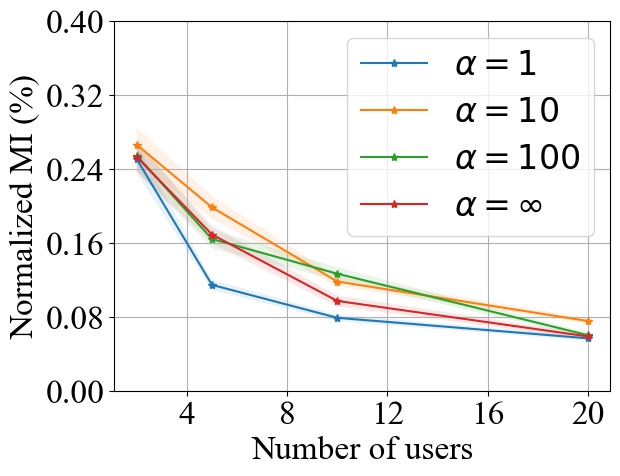}
    \caption{Normalized MI when $E=1$.}
    \label{fig:fedavg_heter1}
\end{subfigure}
\begin{subfigure}{.235\textwidth}
    \includegraphics[width=.99\linewidth]{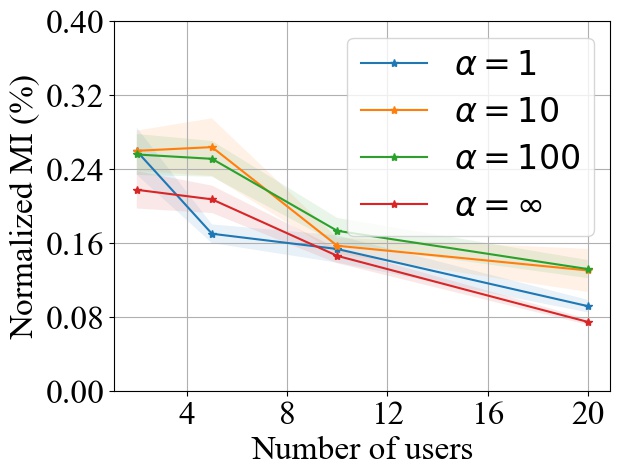}
    \caption{Normalized MI when $E=5$.}
    \label{fig:fedavg_heter2}
\end{subfigure}
\caption{Impact of user heterogeneity when using FedAvg on non-IID CIFAR10. Note that $\alpha=\infty$ means that the user data distributions are identical (IID users), and the MI is normalized by the entropy of a user's local dataset.} 
\label{fig:heter_fedavg}
\vspace{-.1in}
\end{figure}

\begin{figure}[!t]
\centering
    \includegraphics[width=.75\linewidth]{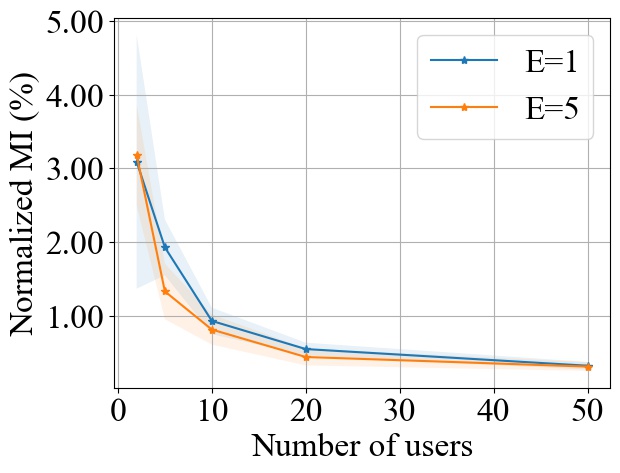}
\caption{Impact of user heterogeneity when using FedAvg on FEMNIST. Note that the MI is normalized by the entropy of target user's local dataset, which is $678 * 176$ .} 
\label{fig:heter_FEMNIST}
\vspace{-.1in}
\end{figure}
%%%%%%%%%%%%%
As discussed in Remark 3 of Section \ref{sec:privacy_guarantee}, in our theoretical analysis, we considered IID data distribution across users in Theorem \ref{Thm:main_Thm} in order to make use of entropic central limit theorem results in developing our upper bounds on privacy leakage. However in practice, the data distribution at the users can be heterogeneous. Hence, in this subsection, we analyze the impact of the non-IID (heterogeneous) data distribution across the users' on the privacy leakage. 
{\color{black}
To measure how user heterogeneity can potentially impact the MI privacy leakage in \texttt{FedAvg}, we  consider two different data settings. In the first setting, we create synthetic users with non-IID data distributions following the methodology in \cite{hsu2019measuring}. For the second setting, we consider FEMNIST \cite{caldas2018leaf}, a benchmark non-IID FL dataset extended from MNIST, which consists of $62$ different classes of 28$\times$28 images ($10$ digits, $26$ lowercase letters, $26$ uppercase letters) written by $3500$ users.

In  the  first, synthetic non-IID data setting, we use Dirichlet distribution parameterized by $\alpha$ to split the dataset into multiple non-IID distributed local datasets. Smaller $\alpha$ (i.e., $\alpha\rightarrow 0$) represents that the users' datasets are more non-identical with each other, while larger $\alpha$ (i.e., $\alpha\rightarrow\infty$)  means that the user datasets are more identical with each other. We choose CIFAR10 as the dataset, CNN as the model, and use \texttt{FedAvg} for a case study while using a batch size of $B=32$. Note that we do not consider \texttt{FedSGD} since it will not be affected by user heterogeneity. During the experiments, we choose the $\alpha$ value from $\{1,10,100,\infty\}$ to create different levels of non-IID user datasets, and we consider $N\in\{2,5,10,20\}$ and $E\in\{1,5\}$.

Fig. \ref{fig:heter_fedavg} shows how the MI privacy leakage varies with the number of users under different $\alpha$, where the MI privacy leakage is normalized by the entropy of each user's local dataset. We notice that the MI privacy leakage will decrease with the number of users consistently under different $\alpha$, which empirically shows that our theoretical results in Section \ref{sec:privacy_guarantee} also holds in the case where users are heterogeneous. 

For the second, FEMNIST data setting, we split the dataset by users into 3500 non-overlapping subsets, each of which contains character images written by a specific user. Considering that the size of each subset is small, in order to have enough training data, we choose to sample $N$ users at each training round instead of using a fixed set of $N$ users, which simulates the user sampling scenario in FL. Specifically, at the beginning of each FL training round with $N$ participating users, we use the same target user and randomly pick the other $N-1$ out of 3500 users.
%Each subset will represent the dataset owned by one federated user out of the $N-1$ selected users.  
%This way of running the experiment not only ensures the data distribution at each round is non-IID, but also simulates the user sampling scenario of FL, where a subset of $N$ users is selected out of $3500$ users in each training round.  
Note that  we consider $N\in\{2,5,10,20,50\}$ and $E\in\{1,5\}$, and use the same model (CNN), batch size ($B=32$), and FedAvg algorithm in our evaluation..

Fig. \ref{fig:heter_FEMNIST} shows  how the MI privacy leakage varies  with the number of users. Similar to the  synthetic  non-IID data setting in Fig. \ref{fig:heter_fedavg}, the privacy leakage decreases with increasing the number of user $N$.
}

\subsection{Practical Privacy Implications}

\subsubsection*{\bf Success of Privacy attacks}
\begin{figure}[!t]
\begin{subfigure}{.235\textwidth}
    \includegraphics[width=.99\linewidth]{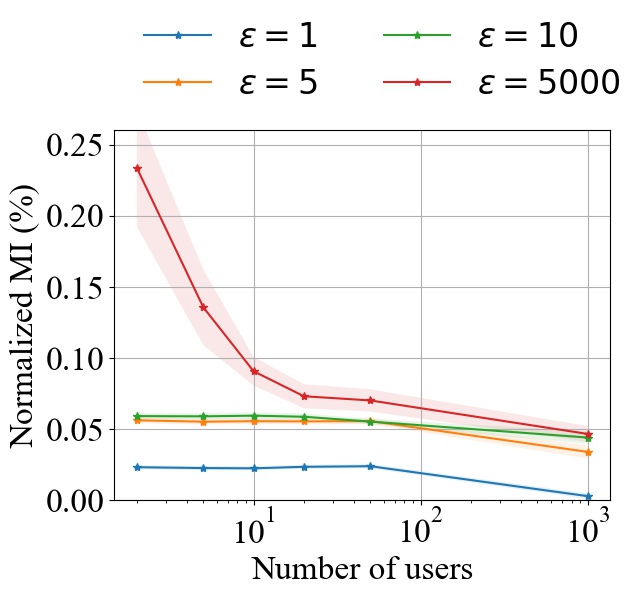}
    \caption{Normalized MI, MNIST.}
    \label{fig:dp_sa_mi}
\end{subfigure}
\begin{subfigure}{.235\textwidth}
    \includegraphics[width=.99\linewidth]{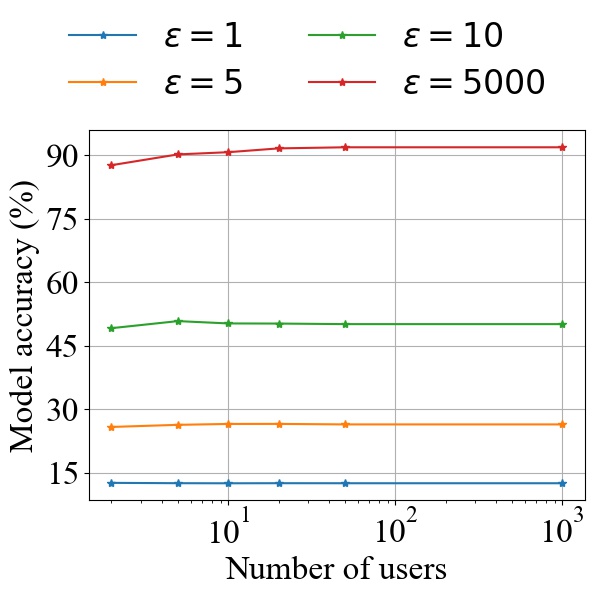}
    \caption{Model accuracy, MNIST.}
    \label{fig:dp_sa_acc}
\end{subfigure}
\caption{Effects of using DP noise together with SA on MI privacy leakage and model accuracy. Note that we add DP noise in aggregated model updates after SA.} 
\label{fig:dp_sa}
\vspace{-.1in}
\end{figure}

{\color{black}To provide insights on how MI translates to practical privacy implications, we conduct experiments using one of the state-of-the-art data reconstruction attack, i.e., the Deep Leakage from Gradients (DLG) attack from \cite{NEURIPS2019_60a6c400}, to show how the MI metric reflects the reconstructed image quality of the attack as we vary system parameters.  Specifically, we choose MNIST as the dataset,  the same SLP used in  Section \ref{Models} as the model, and  FedSGD  with batch size of $32$ as training algorithm. For the data distribution across the users, we consider the IID setting. At the end of each training round, each user uses a batch of images with size 32 to calculate their local gradients, which will be securely aggregated by the server. The DLG attack will reconstruct a batch of images with size 32 from the aggregated gradient, making them as similar as possible to the batch of images used by the target user. After that, we apply the same PSNR (Peak Signal-to-noise Ratio) metric used in \cite{NEURIPS2019_60a6c400} to measure the quality of reconstructed images compared with the images used by the target user during training.  Note that without loss of generality, we report the PSNR value of reconstructed images by DLG attack for the first training round.   
 
Fig. \ref{practical_Implications}  shows the impact of number of users $N$ on the privacy leakage metric (MI) and the reconstructed image quality of DLG attack (PSNR). We pick the image of digit 3 out of the target $32$ images as an example of reconstructed images. We can observe that increasing the number of users $N$ decreases the MI metric as well as the PSNR at almost the  same rate. This demonstrates that  the  MI metric  used in this paper  can translate to practical privacy implications well.

\subsubsection*{\bf MI Privacy leakage under the joint use of DP and SA}
\begin{figure} 
\centering
\includegraphics[width=0.4\textwidth]{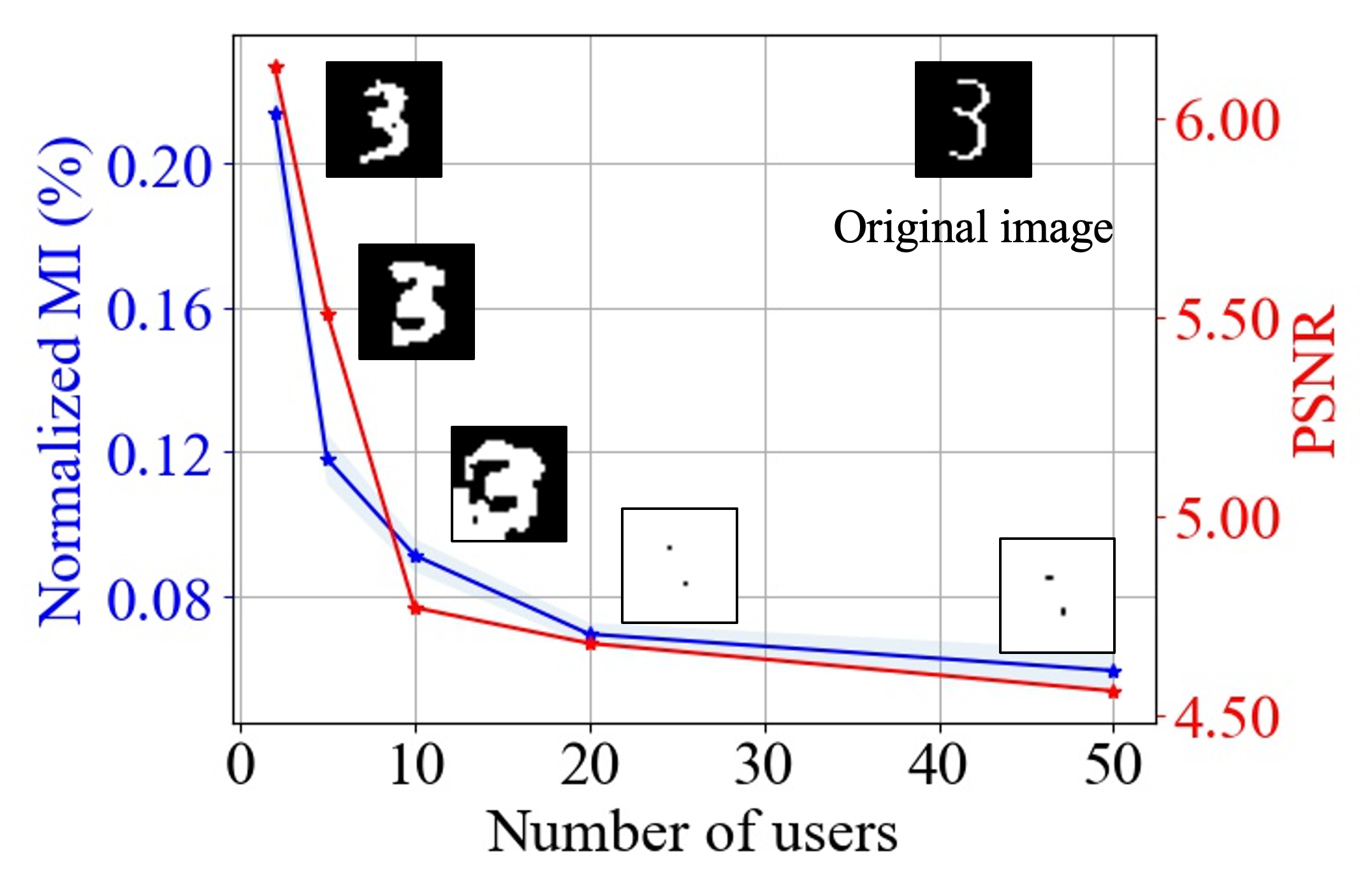}
\caption{Impact of varying the number of users $N$, on the reconstructed image quality (PSNR) of the DLG attack and on the MI privacy leakage.  }
\label{practical_Implications}
\end{figure}

 To highlight the joint effect of differential privacy with secure aggregation, we conduct experiments on the MNIST dataset with a linear model to measure the MI privacy leakage in the presence of centralized DP noise added at the server after SA. Specifically, following~\cite{abadi2016deep}, we first clip the aggregated model updates to make its norm bounded by $C$, and then add Gaussian noise with variance $\sigma^2$ to achieve  $(\epsilon, \delta)$-DP.
 %, in \cite{abadi2016deep}. 
 We set $C=1$, $\delta=1/1200$, and $\sigma=\sqrt{2\log(\frac{1.25}{\delta})}/\epsilon$.

Fig.~\ref{fig:dp_sa_mi} shows the MI privacy leakage for different $(\epsilon, \delta)$-DP levels with SA ($\delta$ is fixed at $1/1200$). As the number of users increase, SA improves the privacy level (measured in terms of MI leakage) for different levels of DP noise, with the effect being most pronounced for weak DP noise level ($\epsilon =5000$ in Fig.~\ref{fig:dp_sa_mi}). 
Our experiments also show that as the number of users increase, the gain from using higher DP noise levels is diminished. In particular, with $N=1000$ users, the MI leakage level for $\epsilon=$5, 10 and 5000  are almost the same; MI leakage is only reduced from 0.046\% to 0.034\% when using $\epsilon=5$ instead of $\epsilon=5000$. In contrast, we get a reduction from 0.234\% to 0.056\% when there are $N=2$ users. 

%This is a very important point so I am putting it in a separate paragraph
Importantly, the reduction observed in privacy leakage due to applying additional DP noise results in a severe degradation in accuracy as seen in Fig.~\ref{fig:dp_sa_acc}, whereas privacy improvement gained by having more users has a negligible effect on the performance of the trained model.
For example, consider the case of 1000 users. One may achieve the same level of privacy in terms of MI leakage (lower than 0.05\% MI) with either (i) $(\epsilon, \delta)$-DP with $\epsilon=10$, which, however, results in unusable model accuracy (less than 50\%), or, (ii) by aggregating the 1000 users and using a tiny amount of DP noise (equivalent to $\epsilon=5000$), which achieves a model accuracy higher than 90\%.
}

\section{Related work}
 {\bf Secure Aggregation in FL.} 
As mentioned secure aggregation has been developed for FL~\cite{cc} to provide protection against model inversion attacks and robustness to user dropouts (due to poor connections or unavailability).
There has been a series of works that aim at improving the efficiency of the aggregation protocol~\cite{secagg_bell2020secure,secagg_so2021securing,secagg_kadhe2020fastsecagg,zhao2021information,so2021lightsecagg,so2021turbo,9712310}. 
This general family of works using secure aggregation disallow the learning information about each client's individual model update beyond the global aggregation of updates, however there has not been a characterization of how much information the global aggregation can leak about the individual client's model and dataset.
To the best of our knowledge, in this work, we provide the first characterization  
% only attempt to characterize the privacy leakage due to the aggregated model through mutual information is by~\cite{yagli2020information}, where they only study a special scenario of distributed mean estimation with datasets sampled from a Gaussian distribution. In contrast, in this work, 
% we characterize 
of the privacy leakage due to the aggregated model through mutual information for FL using secure aggregation.

\noindent{\bf Differential Privacy.}
One way to protect a client's contributions is to use differential privacy (DP). DP provides a rigorous, worst-case mathematical guarantee that the contribution a single client does not impact the result of the query. Central application of differential privacy was studied in~\cite{bassily2014private, chaudhuri2011differentially, abadi2016deep}. This form of central application of DP in FL requires trusting the server with individual model updates before applying the differentially private mechanism. An alternative approach studied in FL for an untrusted server entity is the local differential privacy (LDP) model~\cite{kasiviswanathan2011can,agarwal2018cpsgd,balle2019privacy} were clients apply a differentially private mechanism (e.g. using the Gaussian mechanism) locally on their update before sending to the central server. LDP constraints imply central DP constraints, however due to local privacy constraints LDP mechanisms significantly perturb the input and reduces globally utility due to the compounded effect of adding noise at different clients. 

In this work, we use a mutual information metric to study the privacy guarantees for the client's dataset provided through the secure aggregation protocol without adding differential privacy noise at the clients. In this case, secure aggregation uses contributions from other clients to mask the contribution of a single client. We will discuss in Section~\ref{sec:discussion} situations where relying only on SA can clearly fail to provide differential privacy guarantees and comment on the prevalence of such situations in practical training scenarios.

\noindent{\bf Privacy Attacks.} 
There have been some works  trying to   empirically  show that it is possible to  recovery some  training data from  the  gradient information.  \cite{phong2017privacy, aono2017privacy,wang2019beyond, yin2021gradients}.  Recently, the authors in
\cite{geiping2020inverting} show that it is possible to recover a batch  of   images that were used  in the training  of  
non-smooth deep neural network. In particular, their proposed reconstruction attack was successful in reconstruction of different images from the average  gradient computed over a mini-batch of data.   Their empirical results have shown that the success rate of  the inversion  attack decreases with increasing the batch size.  Similar  observations have  been  demonstrated in  the subsequent works \cite{yin2021gradients}.   In contrast to this work, we are the first to the   best of our knowledge  to theoretically quantify the amount of  information that the aggregated gradient could leak about the private  training data of the users, and to understand how the training   parameters (e.g., number of users) affect the leakage. Additionally, our empirical  results   are different from the ones in \cite{phong2017privacy, aono2017privacy,wang2019beyond, yin2021gradients,yin2021gradients} in   the way of quantifying the leakage. 
In particular, we use the MINE tool to abstractly quantify the amount of  information  leakage in bits instead of the number of the reconstructed images. We have  also empirically  studied  the effect of the system parameters extensively using different  real world data sets and different neural network architectures. 
%\roushdy{TB polished} 
\section{Further Discussion and Conclusions}
\label{sec:discussion}
%%%%%%%%%%%%%%%%%%%
In this paper, we derived the first formal privacy guarantees for FL with SA using MI as a metric to measure how much information the aggregated model update can leak about the local dataset of each user. 
We proved theoretical bounds on the MI privacy leakage in theory and showed through an empirical study that this holds in practice after FL settings. Our concluding observations is that by using FL with SA, we get that: 1) the MI privacy leakage will decrease at a rate of $\mathcal{O}(\frac{1}{N})$ ($N$ is the number of users participating in FL with SA); 2) increasing model size will not have a linear impact on the increase of MI privacy leakage, and the MI privacy leakage only linearly increases with the rank of the covariance matrix of the individual model update; 3) larger batch size during local training can help to reduce the MI privacy leakage. We hope that our findings can shed lights on how to select FL system parameters with SA in practice to reduce privacy leakage and provide an understanding for the baseline protection provided by SA in settings where it is combined with other privacy-preserving approaches such as differential privacy.

\begin{figure}
\centering
\includegraphics[width=0.482\textwidth]{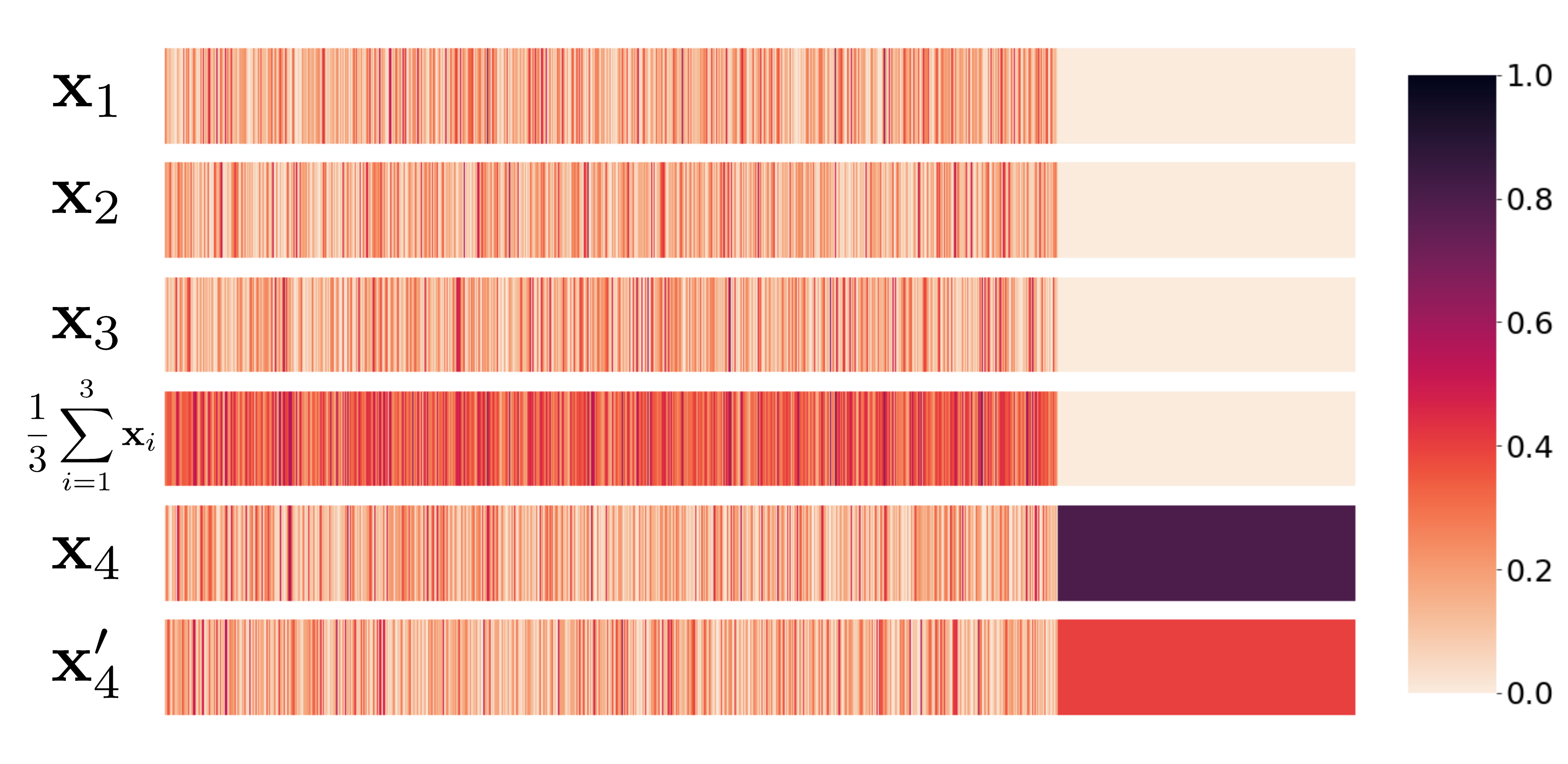}
\vspace{-1em}
\caption{Heatmap of the absolute values of sampled updates from clients $1,2$ and $3$ in the counter example. $\mb{x}_4$ and $\mb{x}_4'$ can be distinguished even adding the aggregated noise from $\sum_{i=1}^3 \mb{x}_i$. }
\label{fig:counter_example}
\end{figure}

\noindent{\bf Can we provide differential privacy guarantees using SA?}
Note that when using FL with SA, then from the point of view of an adversary that is interested in the data of the $i$-th user, the aggregated model in $i^- = [N]\backslash\{i\}$ can be viewed as noise that is independent of the gradient $\mb{x}_i$ given the last global model, which is very similar to an LDP mechanism for the update $\mb{x}_i^{(t)}$ of user $i$ that adds noise to $\mb{x}_i^{(t)}$. This leads to an intriguing question: \textit{Can we get LDP-like guarantees from the securely aggregated updates?}

Since DP is interested in a worst-case guarantee, it turns out that their exist model update distributions where it is impossible to achieve an $\epsilon < \infty$ DP guarantee by using other model updates as noise as illustrated in Fig.~\ref{fig:counter_example}. In this case, the alignment of the sparsity pattern in $x_1, x_2$ and $x_3$ allows an adversary to design a perfect detector to distinguish between $x_4$ and $x_4'$.\\
% We highlight that through our experiments and other results in the literature, we observe that such sparsity alignment happens with very low probability.

\noindent{\bf Why our MI privacy guarantee can avoid this?} Although, the previous example illustrates that DP flavored guarantees are not always possible, in practical scenarios, the worst-case distribution for $\mb{x}_1, \mb{x}_2$ and $\mb{x}_3$ that enables the distinguishing between $\mb{x}_4$ and $\mb{x}_4'$ in Fig.~\ref{fig:counter_example} are an unlikely occurrence during training. For instance, in our theoretical analysis, since users have IID datasets, then having the distribution of $x_1$, $x_2$ and $x_3$ be restricted to a subspace $\mcal{S}_{x_{i^-}}$, implies also that points generated from $\mb{x}_4$ would also belong to $\mcal{S}_{x_{i^-}}$ almost surely. This is a key reason why we can get mutual information guarantee in Theorem ~\ref{Thm:main_Thm}: for an aggregated gradient direction $\sum_{i=1}^N\mb{x}_i$, where each component is restricted to a common subspace $\mcal{S}_x$ protects the contribution of each individual component $\mb{x}_i$ as $N$ increases.

In the worst case, where one component is not restricted to the subspace $\mcal{S}_x$ spanned by the remaining components, then we get the privacy leakage discussed in the example above. We highlight that through our experiments and other studies in the literature~\cite{basak_sparse}, we observe that such sparsity alignment happens with very low probability. This presents motivation for studying a probabilistic notion of DP that satisfies $(\epsilon,\delta)$-DP with a probability at least $\gamma$, instead of the worst-case treatment in current DP notions, but this is beyond the scope of the study in this current work.

Another interesting future direction is to use the results from this work for a providing ``privacy metrics'' to users to estimate/quantify their potential leakage for participating in a federated learning cohort. Such metrics can be embedded in platforms, such as FedML~\cite{he2020fedml}, to guide users to make informed decisions about their participation in federated learning. Finally, it would also be important to extend the results to model aggregation protocols that are beyond weighted averaging (e.g., in federated knowledge transfer~\cite{FedGKT}).
% \subsection{Conclusions}
% In this paper, we perform the first analysis of the formal privacy guarantees for FL with SA. We use MI as a metric to measure how much information the aggregated model update can leak about the local dataset of each user in FL with SA, and then prove the upper bound on such MI privacy leakage in theory. Furthermore, we conduct an empirical study to systematically measure the MI privacy leakage under varieties of FL system steps. Our main findings are 1) the MI privacy leakage will decrease at a rate of $\mathcal{O}(\frac{1}{N})$ ($N$ is the number of users participating in FL with SA); 2) increasing model size will not have a linear impact on the increase of MI privacy leakage, and the MI privacy leakage only linearly increases with the rank of the covariance matrix of the individual model update; 3) larger batch size during local training can help to reduce the MI privacy leakage.  We hope that our findings can shed lights on how to select FL system parameters with SA in practice to reduce privacy leakage.
\bibliographystyle{ACM-Reference-Format}
\bibliography{references2}


\begin{thebibliography}{45}

%%% the Web address.

\ifx \showCODEN    \undefined \def \showCODEN     #1{\unskip}     \fi
\ifx \showDOI      \undefined \def \showDOI       #1{#1}\fi
\ifx \showISBNx    \undefined \def \showISBNx     #1{\unskip}     \fi
\ifx \showISBNxiii \undefined \def \showISBNxiii  #1{\unskip}     \fi
\ifx \showISSN     \undefined \def \showISSN      #1{\unskip}     \fi
\ifx \showLCCN     \undefined \def \showLCCN      #1{\unskip}     \fi
\ifx \shownote     \undefined \def \shownote      #1{#1}          \fi
\ifx \showarticletitle \undefined \def \showarticletitle #1{#1}   \fi
\ifx \showURL      \undefined \def \showURL       {\relax}        \fi
% invisible to TeX
\providecommand\bibfield[2]{#2}
\providecommand\bibinfo[2]{#2}
\providecommand\natexlab[1]{#1}
\providecommand\showeprint[2][]{arXiv:#2}

\bibitem[\protect\citeauthoryear{Abadi, Chu, Goodfellow, McMahan, Mironov,
  Talwar, and Zhang}{Abadi et~al\mbox{.}}{2016}]%
        {abadi2016deep}
\bibfield{author}{\bibinfo{person}{Martin Abadi}, \bibinfo{person}{Andy Chu},
  \bibinfo{person}{Ian Goodfellow}, \bibinfo{person}{H~Brendan McMahan},
  \bibinfo{person}{Ilya Mironov}, \bibinfo{person}{Kunal Talwar}, {and}
  \bibinfo{person}{Li Zhang}.} \bibinfo{year}{2016}\natexlab{}.
\newblock \showarticletitle{Deep learning with differential privacy}. In
  \bibinfo{booktitle}{\emph{Proceedings of the 2016 ACM SIGSAC conference on
  computer and communications security}}. \bibinfo{pages}{308--318}.
\newblock


\bibitem[\protect\citeauthoryear{Agarwal, Suresh, Yu, Kumar, and
  McMahan}{Agarwal et~al\mbox{.}}{2018}]%
        {agarwal2018cpsgd}
\bibfield{author}{\bibinfo{person}{Naman Agarwal},
  \bibinfo{person}{Ananda~Theertha Suresh}, \bibinfo{person}{Felix Xinnan~X
  Yu}, \bibinfo{person}{Sanjiv Kumar}, {and} \bibinfo{person}{Brendan
  McMahan}.} \bibinfo{year}{2018}\natexlab{}.
\newblock \showarticletitle{cpSGD: Communication-efficient and
  differentially-private distributed SGD}.
\newblock \bibinfo{journal}{\emph{Advances in Neural Information Processing
  Systems}}  \bibinfo{volume}{31} (\bibinfo{year}{2018}).
\newblock


\bibitem[\protect\citeauthoryear{Aono, Hayashi, Wang, Moriai,
  et~al\mbox{.}}{Aono et~al\mbox{.}}{2017}]%
        {aono2017privacy}
\bibfield{author}{\bibinfo{person}{Yoshinori Aono}, \bibinfo{person}{Takuya
  Hayashi}, \bibinfo{person}{Lihua Wang}, \bibinfo{person}{Shiho Moriai},
  {et~al\mbox{.}}} \bibinfo{year}{2017}\natexlab{}.
\newblock \showarticletitle{Privacy-preserving deep learning via additively
  homomorphic encryption}.
\newblock \bibinfo{journal}{\emph{IEEE Transactions on Information Forensics
  and Security}} \bibinfo{volume}{13}, \bibinfo{number}{5}
  (\bibinfo{year}{2017}), \bibinfo{pages}{1333--1345}.
\newblock


\bibitem[\protect\citeauthoryear{Balle, Bell, Gasc{\'o}n, and Nissim}{Balle
  et~al\mbox{.}}{2019}]%
        {balle2019privacy}
\bibfield{author}{\bibinfo{person}{Borja Balle}, \bibinfo{person}{James Bell},
  \bibinfo{person}{Adri{\`a} Gasc{\'o}n}, {and} \bibinfo{person}{Kobbi
  Nissim}.} \bibinfo{year}{2019}\natexlab{}.
\newblock \showarticletitle{The privacy blanket of the shuffle model}. In
  \bibinfo{booktitle}{\emph{Annual International Cryptology Conference}}.
  Springer, \bibinfo{pages}{638--667}.
\newblock


\bibitem[\protect\citeauthoryear{Bassily, Smith, and Thakurta}{Bassily
  et~al\mbox{.}}{2014}]%
        {bassily2014private}
\bibfield{author}{\bibinfo{person}{Raef Bassily}, \bibinfo{person}{Adam Smith},
  {and} \bibinfo{person}{Abhradeep Thakurta}.} \bibinfo{year}{2014}\natexlab{}.
\newblock \showarticletitle{Private empirical risk minimization: Efficient
  algorithms and tight error bounds}. In \bibinfo{booktitle}{\emph{2014 IEEE
  55th Annual Symposium on Foundations of Computer Science}}. IEEE,
  \bibinfo{pages}{464--473}.
\newblock


\bibitem[\protect\citeauthoryear{Belghazi, Baratin, Rajeshwar, Ozair, Bengio,
  Courville, and Hjelm}{Belghazi et~al\mbox{.}}{2018}]%
        {belghazi2018mine}
\bibfield{author}{\bibinfo{person}{Mohamed~Ishmael Belghazi},
  \bibinfo{person}{Aristide Baratin}, \bibinfo{person}{Sai Rajeshwar},
  \bibinfo{person}{Sherjil Ozair}, \bibinfo{person}{Yoshua Bengio},
  \bibinfo{person}{Aaron Courville}, {and} \bibinfo{person}{Devon Hjelm}.}
  \bibinfo{year}{2018}\natexlab{}.
\newblock \showarticletitle{Mutual Information Neural Estimation}. In
  \bibinfo{booktitle}{\emph{Proceedings of the 35th International Conference on
  Machine Learning}} \emph{(\bibinfo{series}{Proceedings of Machine Learning
  Research}, Vol.~\bibinfo{volume}{80})},
  \bibfield{editor}{\bibinfo{person}{Jennifer Dy} {and}
  \bibinfo{person}{Andreas Krause}} (Eds.). \bibinfo{publisher}{PMLR},
  \bibinfo{pages}{531--540}.
\newblock


\bibitem[\protect\citeauthoryear{Bell, Bonawitz, Gasc{\'o}n, Lepoint, and
  Raykova}{Bell et~al\mbox{.}}{2020}]%
        {secagg_bell2020secure}
\bibfield{author}{\bibinfo{person}{James~Henry Bell},
  \bibinfo{person}{Kallista~A Bonawitz}, \bibinfo{person}{Adri{\`a}
  Gasc{\'o}n}, \bibinfo{person}{Tancr{\`e}de Lepoint}, {and}
  \bibinfo{person}{Mariana Raykova}.} \bibinfo{year}{2020}\natexlab{}.
\newblock \showarticletitle{Secure single-server aggregation with (poly)
  logarithmic overhead}. In \bibinfo{booktitle}{\emph{Proceedings of the 2020
  ACM SIGSAC Conference on Computer and Communications Security}}.
  \bibinfo{pages}{1253--1269}.
\newblock


\bibitem[\protect\citeauthoryear{Bobkov, Chistyakov, and G{\"o}tze}{Bobkov
  et~al\mbox{.}}{2014}]%
        {bobkov2014berry}
\bibfield{author}{\bibinfo{person}{Sergey~G Bobkov},
  \bibinfo{person}{Gennadiy~P Chistyakov}, {and} \bibinfo{person}{Friedrich
  G{\"o}tze}.} \bibinfo{year}{2014}\natexlab{}.
\newblock \showarticletitle{Berry--Esseen bounds in the entropic central limit
  theorem}.
\newblock \bibinfo{journal}{\emph{Probability Theory and Related Fields}}
  \bibinfo{volume}{159}, \bibinfo{number}{3-4} (\bibinfo{year}{2014}),
  \bibinfo{pages}{435--478}.
\newblock


\bibitem[\protect\citeauthoryear{Bonawitz, Ivanov, Kreuter, Marcedone, McMahan,
  Patel, Ramage, Segal, and Seth}{Bonawitz et~al\mbox{.}}{2017}]%
        {cc}
\bibfield{author}{\bibinfo{person}{Keith Bonawitz}, \bibinfo{person}{Vladimir
  Ivanov}, \bibinfo{person}{Ben Kreuter}, \bibinfo{person}{Antonio Marcedone},
  \bibinfo{person}{H~Brendan McMahan}, \bibinfo{person}{Sarvar Patel},
  \bibinfo{person}{Daniel Ramage}, \bibinfo{person}{Aaron Segal}, {and}
  \bibinfo{person}{Karn Seth}.} \bibinfo{year}{2017}\natexlab{}.
\newblock \showarticletitle{Practical secure aggregation for privacy-preserving
  machine learning}. In \bibinfo{booktitle}{\emph{proceedings of the 2017 ACM
  SIGSAC Conference on Computer and Communications Security}}.
  \bibinfo{pages}{1175--1191}.
\newblock


\bibitem[\protect\citeauthoryear{Caldas, Duddu, Wu, Li, Kone{\v{c}}n{\`y},
  McMahan, Smith, and Talwalkar}{Caldas et~al\mbox{.}}{2018}]%
        {caldas2018leaf}
\bibfield{author}{\bibinfo{person}{Sebastian Caldas}, \bibinfo{person}{Sai
  Meher~Karthik Duddu}, \bibinfo{person}{Peter Wu}, \bibinfo{person}{Tian Li},
  \bibinfo{person}{Jakub Kone{\v{c}}n{\`y}}, \bibinfo{person}{H~Brendan
  McMahan}, \bibinfo{person}{Virginia Smith}, {and} \bibinfo{person}{Ameet
  Talwalkar}.} \bibinfo{year}{2018}\natexlab{}.
\newblock \showarticletitle{Leaf: A benchmark for federated settings}.
\newblock \bibinfo{journal}{\emph{arXiv preprint arXiv:1812.01097}}
  (\bibinfo{year}{2018}).
\newblock


\bibitem[\protect\citeauthoryear{Chaudhuri, Monteleoni, and Sarwate}{Chaudhuri
  et~al\mbox{.}}{2011}]%
        {chaudhuri2011differentially}
\bibfield{author}{\bibinfo{person}{Kamalika Chaudhuri}, \bibinfo{person}{Claire
  Monteleoni}, {and} \bibinfo{person}{Anand~D Sarwate}.}
  \bibinfo{year}{2011}\natexlab{}.
\newblock \showarticletitle{Differentially private empirical risk
  minimization.}
\newblock \bibinfo{journal}{\emph{Journal of Machine Learning Research}}
  \bibinfo{volume}{12}, \bibinfo{number}{3} (\bibinfo{year}{2011}).
\newblock


\bibitem[\protect\citeauthoryear{Cover and Thomas}{Cover and Thomas}{2006}]%
        {10.5555/1146355}
\bibfield{author}{\bibinfo{person}{Thomas~M. Cover} {and}
  \bibinfo{person}{Joy~A. Thomas}.} \bibinfo{year}{2006}\natexlab{}.
\newblock \bibinfo{booktitle}{\emph{Elements of Information Theory (Wiley
  Series in Telecommunications and Signal Processing)}}.
\newblock \bibinfo{publisher}{Wiley-Interscience}, \bibinfo{address}{USA}.
\newblock
\showISBNx{0471241954}


\bibitem[\protect\citeauthoryear{Dong, Chen, Shen, and Wang}{Dong
  et~al\mbox{.}}{2020}]%
        {dong2020eastfly}
\bibfield{author}{\bibinfo{person}{Ye Dong}, \bibinfo{person}{Xiaojun Chen},
  \bibinfo{person}{Liyan Shen}, {and} \bibinfo{person}{Dakui Wang}.}
  \bibinfo{year}{2020}\natexlab{}.
\newblock \showarticletitle{EaSTFLy: Efficient and secure ternary federated
  learning}.
\newblock \bibinfo{journal}{\emph{Computers \& Security}}  \bibinfo{volume}{94}
  (\bibinfo{year}{2020}), \bibinfo{pages}{101824}.
\newblock


\bibitem[\protect\citeauthoryear{Dwork, McSherry, Nissim, and Smith}{Dwork
  et~al\mbox{.}}{2006}]%
        {dwork2006calibrating}
\bibfield{author}{\bibinfo{person}{Cynthia Dwork}, \bibinfo{person}{Frank
  McSherry}, \bibinfo{person}{Kobbi Nissim}, {and} \bibinfo{person}{Adam
  Smith}.} \bibinfo{year}{2006}\natexlab{}.
\newblock \showarticletitle{Calibrating noise to sensitivity in private data
  analysis}. In \bibinfo{booktitle}{\emph{Theory of cryptography conference}}.
  Springer, \bibinfo{pages}{265--284}.
\newblock


\bibitem[\protect\citeauthoryear{Eldan, Mikulincer, and Zhai}{Eldan
  et~al\mbox{.}}{2020}]%
        {eldan2020clt}
\bibfield{author}{\bibinfo{person}{Ronen Eldan}, \bibinfo{person}{Dan
  Mikulincer}, {and} \bibinfo{person}{Alex Zhai}.}
  \bibinfo{year}{2020}\natexlab{}.
\newblock \showarticletitle{The CLT in high dimensions: quantitative bounds via
  martingale embedding}.
\newblock \bibinfo{journal}{\emph{The Annals of Probability}}
  \bibinfo{volume}{48}, \bibinfo{number}{5} (\bibinfo{year}{2020}),
  \bibinfo{pages}{2494--2524}.
\newblock


\bibitem[\protect\citeauthoryear{Elkordy and Salman~Avestimehr}{Elkordy and
  Salman~Avestimehr}{2022}]%
        {9712310}
\bibfield{author}{\bibinfo{person}{Ahmed~Roushdy Elkordy} {and}
  \bibinfo{person}{A. Salman~Avestimehr}.} \bibinfo{year}{2022}\natexlab{}.
\newblock \showarticletitle{HeteroSAg: Secure Aggregation with Heterogeneous
  Quantization in Federated Learning}.
\newblock \bibinfo{journal}{\emph{IEEE Transactions on Communications}}
  (\bibinfo{year}{2022}), \bibinfo{pages}{1--1}.
\newblock
\urldef\tempurl%
\url{https://doi.org/10.1109/TCOMM.2022.3151126}
\showDOI{\tempurl}


\bibitem[\protect\citeauthoryear{Ergun, Sami, and Guler}{Ergun
  et~al\mbox{.}}{2021}]%
        {basak_sparse}
\bibfield{author}{\bibinfo{person}{Irem Ergun}, \bibinfo{person}{Hasin~Us
  Sami}, {and} \bibinfo{person}{Basak Guler}.} \bibinfo{year}{2021}\natexlab{}.
\newblock \showarticletitle{Sparsified Secure Aggregation for
  Privacy-Preserving Federated Learning}.
\newblock \bibinfo{journal}{\emph{arXiv preprint arXiv:2112.12872}}
  (\bibinfo{year}{2021}).
\newblock


\bibitem[\protect\citeauthoryear{Geiping, Bauermeister, Dröge, and
  Moeller}{Geiping et~al\mbox{.}}{2020}]%
        {geiping2020inverting}
\bibfield{author}{\bibinfo{person}{Jonas Geiping}, \bibinfo{person}{Hartmut
  Bauermeister}, \bibinfo{person}{Hannah Dröge}, {and}
  \bibinfo{person}{Michael Moeller}.} \bibinfo{year}{2020}\natexlab{}.
\newblock \showarticletitle{Inverting Gradients -- How easy is it to break
  privacy in federated learning?}. In \bibinfo{booktitle}{\emph{Advances in
  Neural Information Processing Systems}}.
\newblock


\bibitem[\protect\citeauthoryear{He, Annavaram, and Avestimehr}{He
  et~al\mbox{.}}{2020a}]%
        {FedGKT}
\bibfield{author}{\bibinfo{person}{Chaoyang He}, \bibinfo{person}{Murali
  Annavaram}, {and} \bibinfo{person}{Salman Avestimehr}.}
  \bibinfo{year}{2020}\natexlab{a}.
\newblock \showarticletitle{Group knowledge transfer: Federated learning of
  large cnns at the edge}.
\newblock \bibinfo{journal}{\emph{Advances in Neural Information Processing
  Systems}}  \bibinfo{volume}{33} (\bibinfo{year}{2020}),
  \bibinfo{pages}{14068--14080}.
\newblock


\bibitem[\protect\citeauthoryear{He, Li, So, Zeng, Zhang, Wang, Wang,
  Vepakomma, Singh, Qiu, et~al\mbox{.}}{He et~al\mbox{.}}{2020b}]%
        {he2020fedml}
\bibfield{author}{\bibinfo{person}{Chaoyang He}, \bibinfo{person}{Songze Li},
  \bibinfo{person}{Jinhyun So}, \bibinfo{person}{Xiao Zeng},
  \bibinfo{person}{Mi Zhang}, \bibinfo{person}{Hongyi Wang},
  \bibinfo{person}{Xiaoyang Wang}, \bibinfo{person}{Praneeth Vepakomma},
  \bibinfo{person}{Abhishek Singh}, \bibinfo{person}{Hang Qiu},
  {et~al\mbox{.}}} \bibinfo{year}{2020}\natexlab{b}.
\newblock \showarticletitle{Fedml: A research library and benchmark for
  federated machine learning}.
\newblock \bibinfo{journal}{\emph{arXiv preprint arXiv:2007.13518}}
  (\bibinfo{year}{2020}).
\newblock


\bibitem[\protect\citeauthoryear{Hsu, Qi, and Brown}{Hsu et~al\mbox{.}}{2019}]%
        {hsu2019measuring}
\bibfield{author}{\bibinfo{person}{Tzu-Ming~Harry Hsu}, \bibinfo{person}{Hang
  Qi}, {and} \bibinfo{person}{Matthew Brown}.} \bibinfo{year}{2019}\natexlab{}.
\newblock \showarticletitle{Measuring the effects of non-identical data
  distribution for federated visual classification}.
\newblock \bibinfo{journal}{\emph{arXiv preprint arXiv:1909.06335}}
  (\bibinfo{year}{2019}).
\newblock


\bibitem[\protect\citeauthoryear{Kadhe, Rajaraman, Koyluoglu, and
  Ramchandran}{Kadhe et~al\mbox{.}}{2020}]%
        {secagg_kadhe2020fastsecagg}
\bibfield{author}{\bibinfo{person}{Swanand Kadhe}, \bibinfo{person}{Nived
  Rajaraman}, \bibinfo{person}{O~Ozan Koyluoglu}, {and} \bibinfo{person}{Kannan
  Ramchandran}.} \bibinfo{year}{2020}\natexlab{}.
\newblock \showarticletitle{Fastsecagg: Scalable secure aggregation for
  privacy-preserving federated learning}.
\newblock \bibinfo{journal}{\emph{arXiv preprint arXiv:2009.11248}}
  (\bibinfo{year}{2020}).
\newblock


\bibitem[\protect\citeauthoryear{Kairouz, Liu, and Steinke}{Kairouz
  et~al\mbox{.}}{2021}]%
        {kairouz2021distributed}
\bibfield{author}{\bibinfo{person}{Peter Kairouz}, \bibinfo{person}{Ziyu Liu},
  {and} \bibinfo{person}{Thomas Steinke}.} \bibinfo{year}{2021}\natexlab{}.
\newblock \showarticletitle{The distributed discrete gaussian mechanism for
  federated learning with secure aggregation}.
\newblock \bibinfo{journal}{\emph{arXiv preprint arXiv:2102.06387}}
  (\bibinfo{year}{2021}).
\newblock


\bibitem[\protect\citeauthoryear{Kairouz, McMahan, Brendan, and et~al.}{Kairouz
  et~al\mbox{.}}{2019}]%
        {kairouz2019advances}
\bibfield{author}{\bibinfo{person}{Peter Kairouz}, \bibinfo{person}{H.~Brendan
  McMahan}, \bibinfo{person}{Brendan}, {and} \bibinfo{person}{et al.}}
  \bibinfo{year}{2019}\natexlab{}.
\newblock \showarticletitle{Advances and Open Problems in Federated Learning}.
\newblock \bibinfo{journal}{\emph{preprint arXiv:1912.04977}}
  (\bibinfo{year}{2019}).
\newblock
\showeprint[arxiv]{1912.04977}


\bibitem[\protect\citeauthoryear{Kasiviswanathan, Lee, Nissim, Raskhodnikova,
  and Smith}{Kasiviswanathan et~al\mbox{.}}{2011}]%
        {kasiviswanathan2011can}
\bibfield{author}{\bibinfo{person}{Shiva~Prasad Kasiviswanathan},
  \bibinfo{person}{Homin~K Lee}, \bibinfo{person}{Kobbi Nissim},
  \bibinfo{person}{Sofya Raskhodnikova}, {and} \bibinfo{person}{Adam Smith}.}
  \bibinfo{year}{2011}\natexlab{}.
\newblock \showarticletitle{What can we learn privately?}
\newblock \bibinfo{journal}{\emph{SIAM J. Comput.}} \bibinfo{volume}{40},
  \bibinfo{number}{3} (\bibinfo{year}{2011}), \bibinfo{pages}{793--826}.
\newblock


\bibitem[\protect\citeauthoryear{Krizhevsky}{Krizhevsky}{2009}]%
        {krizhevsky2009learning}
\bibfield{author}{\bibinfo{person}{Alex Krizhevsky}.}
  \bibinfo{year}{2009}\natexlab{}.
\newblock \bibinfo{booktitle}{\emph{Learning multiple layers of features from
  tiny images}}.
\newblock \bibinfo{type}{{T}echnical {R}eport}.
  \bibinfo{institution}{Citeseer}.
\newblock


\bibitem[\protect\citeauthoryear{Krizhevsky, Sutskever, and Hinton}{Krizhevsky
  et~al\mbox{.}}{2012}]%
        {krizhevsky2012imagenet}
\bibfield{author}{\bibinfo{person}{Alex Krizhevsky}, \bibinfo{person}{Ilya
  Sutskever}, {and} \bibinfo{person}{Geoffrey~E Hinton}.}
  \bibinfo{year}{2012}\natexlab{}.
\newblock \showarticletitle{Imagenet classification with deep convolutional
  neural networks}.
\newblock \bibinfo{journal}{\emph{Advances in neural information processing
  systems}}  \bibinfo{volume}{25} (\bibinfo{year}{2012}).
\newblock


\bibitem[\protect\citeauthoryear{Kuznetsov, Chen, and Zhao}{Kuznetsov
  et~al\mbox{.}}{2021}]%
        {9708971}
\bibfield{author}{\bibinfo{person}{Eugene Kuznetsov}, \bibinfo{person}{Yitao
  Chen}, {and} \bibinfo{person}{Ming Zhao}.} \bibinfo{year}{2021}\natexlab{}.
\newblock \showarticletitle{SecureFL: Privacy Preserving Federated Learning
  with SGX and TrustZone}. In \bibinfo{booktitle}{\emph{2021 IEEE/ACM Symposium
  on Edge Computing (SEC)}}. \bibinfo{pages}{55--67}.
\newblock
\urldef\tempurl%
\url{https://doi.org/10.1145/3453142.3491287}
\showDOI{\tempurl}


\bibitem[\protect\citeauthoryear{LeCun, Cortes, and Burges}{LeCun
  et~al\mbox{.}}{2010}]%
        {MNIST}
\bibfield{author}{\bibinfo{person}{Yann LeCun}, \bibinfo{person}{Corinna
  Cortes}, {and} \bibinfo{person}{CJ Burges}.} \bibinfo{year}{2010}\natexlab{}.
\newblock \showarticletitle{MNIST handwritten digit database}.
\newblock \bibinfo{journal}{\emph{ATT Labs [Online]. Available:
  http://yann.lecun.com/exdb/mnist}}  \bibinfo{volume}{2}
  (\bibinfo{year}{2010}).
\newblock


\bibitem[\protect\citeauthoryear{McMahan, Moore, Ramage, Hampson, and
  y~Arcas}{McMahan et~al\mbox{.}}{2017}]%
        {FedAvg}
\bibfield{author}{\bibinfo{person}{Brendan McMahan}, \bibinfo{person}{Eider
  Moore}, \bibinfo{person}{Daniel Ramage}, \bibinfo{person}{Seth Hampson},
  {and} \bibinfo{person}{Blaise~Aguera y Arcas}.}
  \bibinfo{year}{2017}\natexlab{}.
\newblock \showarticletitle{{Communication-Efficient Learning of Deep Networks
  from Decentralized Data}}. In \bibinfo{booktitle}{\emph{Proceedings of the
  20th International Conference on Artificial Intelligence and Statistics}}
  \emph{(\bibinfo{series}{Proceedings of Machine Learning Research},
  Vol.~\bibinfo{volume}{54})}, \bibfield{editor}{\bibinfo{person}{Aarti Singh}
  {and} \bibinfo{person}{Jerry Zhu}} (Eds.). \bibinfo{pages}{1273--1282}.
\newblock


\bibitem[\protect\citeauthoryear{Mugunthan, Polychroniadou, Byrd, and
  Balch}{Mugunthan et~al\mbox{.}}{2019}]%
        {mugunthan2019smpai}
\bibfield{author}{\bibinfo{person}{Vaikkunth Mugunthan},
  \bibinfo{person}{Antigoni Polychroniadou}, \bibinfo{person}{David Byrd},
  {and} \bibinfo{person}{Tucker~Hybinette Balch}.}
  \bibinfo{year}{2019}\natexlab{}.
\newblock \showarticletitle{Smpai: Secure multi-party computation for federated
  learning}. In \bibinfo{booktitle}{\emph{Proceedings of the NeurIPS 2019
  Workshop on Robust AI in Financial Services}}.
\newblock


\bibitem[\protect\citeauthoryear{Phong, Aono, Hayashi, Wang, and Moriai}{Phong
  et~al\mbox{.}}{2017}]%
        {phong2017privacy}
\bibfield{author}{\bibinfo{person}{Le~Trieu Phong}, \bibinfo{person}{Yoshinori
  Aono}, \bibinfo{person}{Takuya Hayashi}, \bibinfo{person}{Lihua Wang}, {and}
  \bibinfo{person}{Shiho Moriai}.} \bibinfo{year}{2017}\natexlab{}.
\newblock \showarticletitle{Privacy-preserving deep learning: Revisited and
  enhanced}. In \bibinfo{booktitle}{\emph{International Conference on
  Applications and Techniques in Information Security}}. Springer,
  \bibinfo{pages}{100--110}.
\newblock


\bibitem[\protect\citeauthoryear{Sahu, Li, Sanjabi, Zaheer, Talwalkar, and
  Smith}{Sahu et~al\mbox{.}}{2018}]%
        {fedprox_paper}
\bibfield{author}{\bibinfo{person}{Anit~Kumar Sahu}, \bibinfo{person}{Tian Li},
  \bibinfo{person}{Maziar Sanjabi}, \bibinfo{person}{Manzil Zaheer},
  \bibinfo{person}{Ameet Talwalkar}, {and} \bibinfo{person}{Virginia Smith}.}
  \bibinfo{year}{2018}\natexlab{}.
\newblock \showarticletitle{On the convergence of federated optimization in
  heterogeneous networks}.
\newblock \bibinfo{journal}{\emph{arXiv preprint arXiv:1812.06127}}
  \bibinfo{volume}{3} (\bibinfo{year}{2018}), \bibinfo{pages}{3}.
\newblock


\bibitem[\protect\citeauthoryear{Simsekli, Sagun, and Gurbuzbalaban}{Simsekli
  et~al\mbox{.}}{2019}]%
        {alpha_stable_noise_SGD}
\bibfield{author}{\bibinfo{person}{Umut Simsekli}, \bibinfo{person}{Levent
  Sagun}, {and} \bibinfo{person}{Mert Gurbuzbalaban}.}
  \bibinfo{year}{2019}\natexlab{}.
\newblock \showarticletitle{A tail-index analysis of stochastic gradient noise
  in deep neural networks}. In \bibinfo{booktitle}{\emph{International
  Conference on Machine Learning}}. PMLR, \bibinfo{pages}{5827--5837}.
\newblock


\bibitem[\protect\citeauthoryear{So, Ali, Guler, Jiao, and Avestimehr}{So
  et~al\mbox{.}}{2021a}]%
        {secagg_so2021securing}
\bibfield{author}{\bibinfo{person}{Jinhyun So}, \bibinfo{person}{Ramy~E Ali},
  \bibinfo{person}{Basak Guler}, \bibinfo{person}{Jiantao Jiao}, {and}
  \bibinfo{person}{Salman Avestimehr}.} \bibinfo{year}{2021}\natexlab{a}.
\newblock \showarticletitle{Securing secure aggregation: Mitigating multi-round
  privacy leakage in federated learning}.
\newblock \bibinfo{journal}{\emph{arXiv preprint arXiv:2106.03328}}
  (\bibinfo{year}{2021}).
\newblock


\bibitem[\protect\citeauthoryear{So, G{\"u}ler, and Avestimehr}{So
  et~al\mbox{.}}{2021b}]%
        {so2021turbo}
\bibfield{author}{\bibinfo{person}{Jinhyun So}, \bibinfo{person}{Ba{\c{s}}ak
  G{\"u}ler}, {and} \bibinfo{person}{A~Salman Avestimehr}.}
  \bibinfo{year}{2021}\natexlab{b}.
\newblock \showarticletitle{Turbo-aggregate: Breaking the quadratic aggregation
  barrier in secure federated learning}.
\newblock \bibinfo{journal}{\emph{IEEE Journal on Selected Areas in Information
  Theory}} \bibinfo{volume}{2}, \bibinfo{number}{1} (\bibinfo{year}{2021}),
  \bibinfo{pages}{479--489}.
\newblock


\bibitem[\protect\citeauthoryear{So, Nolet, Yang, Li, Yu, E~Ali, Guler, and
  Avestimehr}{So et~al\mbox{.}}{2022}]%
        {so2021lightsecagg}
\bibfield{author}{\bibinfo{person}{Jinhyun So}, \bibinfo{person}{Corey~J
  Nolet}, \bibinfo{person}{Chien-Sheng Yang}, \bibinfo{person}{Songze Li},
  \bibinfo{person}{Qian Yu}, \bibinfo{person}{Ramy E~Ali},
  \bibinfo{person}{Basak Guler}, {and} \bibinfo{person}{Salman Avestimehr}.}
  \bibinfo{year}{2022}\natexlab{}.
\newblock \showarticletitle{Lightsecagg: a lightweight and versatile design for
  secure aggregation in federated learning}.
\newblock \bibinfo{journal}{\emph{Proceedings of Machine Learning and Systems}}
   \bibinfo{volume}{4} (\bibinfo{year}{2022}), \bibinfo{pages}{694--720}.
\newblock


\bibitem[\protect\citeauthoryear{Truex, Baracaldo, Anwar, Steinke, Ludwig,
  Zhang, and Zhou}{Truex et~al\mbox{.}}{2019}]%
        {truex2019hybrid}
\bibfield{author}{\bibinfo{person}{Stacey Truex}, \bibinfo{person}{Nathalie
  Baracaldo}, \bibinfo{person}{Ali Anwar}, \bibinfo{person}{Thomas Steinke},
  \bibinfo{person}{Heiko Ludwig}, \bibinfo{person}{Rui Zhang}, {and}
  \bibinfo{person}{Yi Zhou}.} \bibinfo{year}{2019}\natexlab{}.
\newblock \showarticletitle{A hybrid approach to privacy-preserving federated
  learning}. In \bibinfo{booktitle}{\emph{Proceedings of the 12th ACM Workshop
  on Artificial Intelligence and Security}}. \bibinfo{pages}{1--11}.
\newblock


\bibitem[\protect\citeauthoryear{Wang, Song, Zhang, Song, Wang, and Qi}{Wang
  et~al\mbox{.}}{2019}]%
        {wang2019beyond}
\bibfield{author}{\bibinfo{person}{Zhibo Wang}, \bibinfo{person}{Mengkai Song},
  \bibinfo{person}{Zhifei Zhang}, \bibinfo{person}{Yang Song},
  \bibinfo{person}{Qian Wang}, {and} \bibinfo{person}{Hairong Qi}.}
  \bibinfo{year}{2019}\natexlab{}.
\newblock \showarticletitle{Beyond inferring class representatives: User-level
  privacy leakage from federated learning}. In \bibinfo{booktitle}{\emph{IEEE
  INFOCOM 2019-IEEE Conference on Computer Communications}}. IEEE,
  \bibinfo{pages}{2512--2520}.
\newblock


\bibitem[\protect\citeauthoryear{Xu, Baracaldo, Zhou, Anwar, and Ludwig}{Xu
  et~al\mbox{.}}{2019}]%
        {xu2019hybridalpha}
\bibfield{author}{\bibinfo{person}{Runhua Xu}, \bibinfo{person}{Nathalie
  Baracaldo}, \bibinfo{person}{Yi Zhou}, \bibinfo{person}{Ali Anwar}, {and}
  \bibinfo{person}{Heiko Ludwig}.} \bibinfo{year}{2019}\natexlab{}.
\newblock \showarticletitle{Hybridalpha: An efficient approach for
  privacy-preserving federated learning}. In
  \bibinfo{booktitle}{\emph{Proceedings of the 12th ACM Workshop on Artificial
  Intelligence and Security}}. \bibinfo{pages}{13--23}.
\newblock


\bibitem[\protect\citeauthoryear{Yin, Mallya, Vahdat, Alvarez, Kautz, and
  Molchanov}{Yin et~al\mbox{.}}{2021}]%
        {yin2021gradients}
\bibfield{author}{\bibinfo{person}{Hongxu Yin}, \bibinfo{person}{Arun Mallya},
  \bibinfo{person}{Arash Vahdat}, \bibinfo{person}{Jose~M. Alvarez},
  \bibinfo{person}{Jan Kautz}, {and} \bibinfo{person}{Pavlo Molchanov}.}
  \bibinfo{year}{2021}\natexlab{}.
\newblock \showarticletitle{See through Gradients: Image Batch Recovery via
  GradInversion}.
\newblock \bibinfo{journal}{\emph{arXiv,2104.07586}} (\bibinfo{year}{2021}).
\newblock


\bibitem[\protect\citeauthoryear{Zhang, Wang, Cao, Hou, and Meng}{Zhang
  et~al\mbox{.}}{2021}]%
        {zhang2021shufflefl}
\bibfield{author}{\bibinfo{person}{Yuhui Zhang}, \bibinfo{person}{Zhiwei Wang},
  \bibinfo{person}{Jiangfeng Cao}, \bibinfo{person}{Rui Hou}, {and}
  \bibinfo{person}{Dan Meng}.} \bibinfo{year}{2021}\natexlab{}.
\newblock \showarticletitle{ShuffleFL: gradient-preserving federated learning
  using trusted execution environment}. In
  \bibinfo{booktitle}{\emph{Proceedings of the 18th ACM International
  Conference on Computing Frontiers}}. \bibinfo{pages}{161--168}.
\newblock


\bibitem[\protect\citeauthoryear{Zhao and Sun}{Zhao and Sun}{2021}]%
        {zhao2021information}
\bibfield{author}{\bibinfo{person}{Yizhou Zhao} {and} \bibinfo{person}{Hua
  Sun}.} \bibinfo{year}{2021}\natexlab{}.
\newblock \showarticletitle{Information theoretic secure aggregation with user
  dropouts}. In \bibinfo{booktitle}{\emph{2021 IEEE International Symposium on
  Information Theory (ISIT)}}. IEEE, \bibinfo{pages}{1124--1129}.
\newblock


\bibitem[\protect\citeauthoryear{Zhu, Liu, and Han}{Zhu et~al\mbox{.}}{2019a}]%
        {NEURIPS2019_60a6c400}
\bibfield{author}{\bibinfo{person}{Ligeng Zhu}, \bibinfo{person}{Zhijian Liu},
  {and} \bibinfo{person}{Song Han}.} \bibinfo{year}{2019}\natexlab{a}.
\newblock \showarticletitle{Deep Leakage from Gradients}. In
  \bibinfo{booktitle}{\emph{Advances in Neural Information Processing
  Systems}}, Vol.~\bibinfo{volume}{32}.
\newblock


\bibitem[\protect\citeauthoryear{Zhu, Wu, Yu, Wu, and Ma}{Zhu
  et~al\mbox{.}}{2019b}]%
        {anisotropic_noise_SGD}
\bibfield{author}{\bibinfo{person}{Zhanxing Zhu}, \bibinfo{person}{Jingfeng
  Wu}, \bibinfo{person}{Bing Yu}, \bibinfo{person}{Lei Wu}, {and}
  \bibinfo{person}{Jinwen Ma}.} \bibinfo{year}{2019}\natexlab{b}.
\newblock \showarticletitle{The Anisotropic Noise in Stochastic Gradient
  Descent: Its Behavior of Escaping from Sharp Minima and Regularization
  Effects}. In \bibinfo{booktitle}{\emph{International Conference on Machine
  Learning}}. PMLR, \bibinfo{pages}{7654--7663}.
\newblock


\end{thebibliography}

\appendix
%\newpage
% \onecolumn
%\newpage
% \section{Appendix}
% \label{sec:appendix}
\section{Proof of Theorem 1}\label{append:proof_theorem}
% \noindent \textbf{Proof of Theorem 1} \\
Without loss of generality, using permutation of clients indices, we  will prove the upper bound  for the following term 
\begin{equation} \label{eq-40}
I\left(\mathbf{x}^{(t)}_N ;   \frac{1}{N}  \sum_{i=1}^N \mathbf{x}^{(t)}_i\middle|  \left \{ \frac{1}{N}  \sum_{i =1  }^N \mathbf{x}^{(k)}_i  \right\}_{k \in [t-1]}  \right), 
\end{equation}
where $\mathbf{x}_N$ is  the mini-batch gradient of node $i$   which  is  given by
\begin{equation}\label{eq-3}
\mathbf {x}_i^{(t)} = \frac{1}{B} \sum_{b \in \mathcal{B}^{(t)}_i } g_i( \bm {\theta}^{(t)}, b),
\end{equation}

\noindent We will use the following property of vectors with singular covariance matrices in the proof of  this theorem.

\begin{prope}\label{prop:nonsingular_to_singular}
\textit{Given a random vector $\mathbf{q}$ with a  singular covariance matrix $\mathbf{K}_q$ of rank $d^*$, there exists a sub-vector $\bar{\mathbf{q}}$ of $\mathbf{q}$ with a non-singular covariance matrix $\mathbf{K}_{\bar{q}}$ such that $\mathbf{q}[k] = \mb{A}\bar{\mathbf{q}}$ where $\mb{A} \in \mbb{R}^{d\times d^*}$ is a deterministic linear transformation matrix.}
\end{prope}

Let us define $S_{{N}}^{(t)} = \frac{1}{{N}}\sum_{i = 1}^N 
\mathbf{x}_i^{(t)}$. We also use the  the definition   of  $\bar{{g}}_i( \bm {\theta}^{(t)}, b) \in \mathbb{R}^{d^*}$, for $d^* \leq d$ where $d$ is the model size,   which is  the largest sub-vector of the  stochastic gradient   $ g_i( \bm {\theta}^{(t)}, b) $  such that $\bar{{g}}_i( \bm {\theta}^{(t)}, b)$ has a non-singular covariance matrix  $K_{\bar{G}^{(t)}}$ for all $ i \in \mathcal{N}$. According to the definition of   $\bar{{g}}_i( \bm {\theta}^{(t)}, b)$, we can rewrite \eqref{eq-40}  and the term $S_{{N}}^{(t)}$ as follows:

\begin{align}
    \bar{\mathbf {x}}_i^{(t)} &= \frac{1}{B} \sum_{b \in \mathcal{D}_i } \bar{{g}}_i( \bm {\theta}^{(t)}, b)
    \nonumber \\  \bar{S}_{{N}}^{(t)}&=  \frac{1}{{N}}\sum_{i \in \mathcal{N}}
\bar{\mathbf{x}}_i^{(t)}
\end{align}

 Let also define  ${F}_{{N}}^{(t)} = \sqrt{N} \bar{S}_{{N}}^{(t)}$. We can decompose  the expression in  \eqref{eq-40} as follows:
\begin{align}\label{eq:mutual_info_decomp_1}
     &I\left(\mathbf{x}^{(t)}_N ;   S_{{N}}^{(t)}\middle|  \left\{S_{{N}}^{(k)} \right\}_{k\in[t-1]} \right) \nonumber \\ &\overset{(a)}= I\left(\sqrt{B}\mathbf{x}_N^{(t)}; \sqrt{N}S_{{N}}^{(t)}\middle| \left\{S_{{N}}^{(k)} \right\}_{k\in[t-1]} \right) 
    \nonumber \\
    &\overset{(b)} = I\left(\sqrt{B}\bar{\mathbf{x}}_N^{(t)}; F_{{N}}^{(t)} \middle| \left\{S_{{N}}^{(k)} \right\}_{k\in[t-1]} \right) \nonumber \\
   & = h\left(\!\!\!\sqrt{B}\bar{\mathbf{x}}_N^{(t)}\!\middle|\! \left\{\!S_{{N}}^{(k)} \!\right\}_{\!k\in[t-1]}\!\right) {+} h\left(\!\!{F}_{{N}}^{(t)}\!\middle|\!\left\{S_{{N}}^{(k)} \!\right\}_{\!k\in[t-1]}\! \right)\nonumber\\
   &\qquad- h\left(\sqrt{B}\bar{\mathbf{x}}_N^{(t)}, F_{{N}}^{(t)} \middle| \left\{S_{{N}}^{(k)} \right\}_{k\in[t-1]} \right) \nonumber\\
    &= h\left(\!\!\!\sqrt{B}\bar{\mathbf{x}}_N^{(t)}\!\middle|\! \left\{\!S_{{N}}^{(k)} \!\right\}_{\!k\in[t-1]}\!\right) {+} h\left(\!\!{F}_{{N}}^{(t)}\!\middle|\!\left\{S_{{N}}^{(k)} \!\right\}_{\!k\in[t-1]}\! \right)\nonumber\\
    &\qquad-\! h\left(\!\begin{bmatrix} \mathrm{I}_d^* & \mathrm{0}_d^* \nonumber\\ \mathrm{I}_d^*\frac{1}{\sqrt{N}} & \frac{\sqrt{N-1}}{\sqrt{N}}\mathrm{I}_d^*\end{bmatrix}\!\!\!\begin{bmatrix} \sqrt{B}\bar{\mathbf{x}}_N^{(t)} \\ {F}_{{N-1}}^{(t)}\end{bmatrix}\! \middle|\! \left\{S_{{N}}^{(k)} \right\}_{\!k\in[t-1]}\!\right) \nonumber\\
    &\stackrel{(c)}{=} h\left(\!\!\!\sqrt{B}\bar{\mathbf{x}}_N^{(t)}\!\middle|\! \left\{\!S_{{N}}^{(k)} \!\right\}_{\!k\in[t-1]}\!\right) {+} h\left(\!\!{F}_{{N}}^{(t)}\!\middle|\!\left\{S_{{N}}^{(k)} \!\right\}_{\!k\in[t-1]}\!\right)\nonumber\\
    &\qquad{-} h\!\left(\!\!\!\sqrt{B}\bar{\mathbf{x}}_N^{(t)}\!\middle|\! \left\{\!S_{{N}}^{(k)} \!\right\}_{\!\!k\in[t-1]}\!\right)\!{-}h\!\left(\!{F}_{{N-1}}^{(t)}\!\middle|\! \left\{S_{{N}}^{(k)} \right\}_{\!\!k\in[t-1]}\!\right)\nonumber\\
    &\qquad- \log\left|\det \begin{bmatrix} \mathrm{I}_d^* & \mathrm{0}_d^* \\ \mathrm{I}_d^*\frac{1}{\sqrt{N-1}} & \frac{\sqrt{N-1}}{\sqrt{N}}\mathrm{I}_d^*\end{bmatrix} \right|\nonumber \\
    &\stackrel{(d)}= h\left(\!F_{{N}}^{(t)}\middle| \left\{S_{{N}}^{(k)} \right\}_{\!k\in[t-1]}\!\right)\nonumber\\
    &\qquad- h\!\left(\!\!{F}_{{N-1}}^{(t)}\middle|\! \left\{S_{{N}}^{(k)} \right\}_{\!k\in[t-1]}\!\right) {+} \frac{d^*}{2}\log\left(\!\frac{N}{N-1}\!\right)\!,
\end{align}

where: $(a)$ follows from the fact that the mutual information is invariant under deterministic multiplication; $(b)$ from Property~\ref{prop:nonsingular_to_singular}  $(c)$ follows from the property of the entropy of linear transformation of random vectors \cite{10.5555/1146355} and the fact that $\bar{\mathbf{x}}_N^{(t)}$  and ${F}_{{N-1}}^{(t)}$ are conditionally independent given $\left\{S_{{N}}^{(k)} \right\}_{k\in[t-1]}$ (e.g., the last global model at time $t$) ; $(d)$ follows from the Schur compliment of the matrix.

We will now turn our attention to characterizing the entropy term $h\left(F_{M}^{(t)}\middle| \left\{S_{{N}}^{(k)} \right\}_{k\in[t-1]}\right)$ for any ${M}$. Note that

\begin{align}\label{eq:mutual_info_decomp_2}
    &h\left(F_{M}^{(t)}\middle| \left\{S_{{N}}^{(k)} \right\}_{k\in[t-1]}\right) \nonumber \\ 
    &= h \left(\!\!\frac{1}{\sqrt{MB}} \sum_{i=1}^M  \sum_{d \in \mathcal{B}_i^{(t)}} \bar{g}_i(b, \bm {\theta}^{(t)}) \middle| \left\{S_{{N}}^{(k)} \right\}_{k\in[t-1]}\!\!\right) \nonumber \\
    &\stackrel{(i)}{=} h\!\! \left(\frac{\mathbf{K}_{\bar{G}^{(t)}}^{1/2}}{\sqrt{MB}} \sum_{i=1}^M\!\sum_{b \in \mathcal{B}_i^{(t)}} \!\!\!\widehat{g}_i(b, \bm {\theta}^{(t)})\middle| \left\{S_{{N}}^{(k)} \right\}_{k\in[t-1]}\right) \nonumber \\
    &\stackrel{(ii)}= \log\left(|\det \mathbf{K}_{\bar{G}^{(t)}}^{1/2} |\right) \nonumber \\ 
    &\qquad {+} \underbrace{h\!\!\left(\!\!\frac{1}{\sqrt{MB}}\!\sum_{i=1}^M\!\sum_{b \in \mathcal{B}_i^{(t)}} \!\!\!\widehat{g}_i(b, \bm {\theta}^{(t)})\middle|\! \left\{S_{{N}}^{(k)} \right\}_{\!k\in[t-1]}\!\right)}_{H_M}\!,
\end{align}
where: $(i)$ makes use of the fact that the covariance matrix is the same across clients and  using the   whitening definition  (Definition \ref{def-2}) on the vector  $\bar{g}_i(b, \bm {\theta}^{(t)})$; $(ii)$ again uses the property of entropy of linear transformation of random vectors.

Note that the term of $h\left(F_{M}^{(t)}\middle| \left\{S_{{N}}^{(k)} \right\}_{k\in[t-1]}\right)$ only depends on $M$ in the second term $H_M$. As a result by substituting~\eqref{eq:mutual_info_decomp_2} in~\eqref{eq:mutual_info_decomp_1}, we get that
\begin{align}\label{eq:mutual_info_restated}
   & I\left(\mathbf{x}^{(t)}_N ;   S_{{N}}^{(t)}\middle|  \left\{S_{{N}}^{(k)} \right\}_{k\in[t-1]} \right)\nonumber \\
   & = H_N - H_{N-1} + \frac{d^*}{2}\log\left(\frac{N}{N-1}\right),
\end{align}
Our final step is to find suitable upper and lower bounds for $H_M$ to use in~\eqref{eq:mutual_info_restated}. Recall for the following arguments that due to whitening, the vector $\widehat{g}_b^{(t)} = \widehat{g}(b, \bm {\theta}^{(t)})$ has zero mean and identity covariance.

\subsection{Upper bound on \texorpdfstring{$H_M$}{H\textunderscore M}} 
The upper bound is the simplest due to basic entropy properties. In particular, the sum $\frac{1}{\sqrt{MB}} \sum_{i=1}^M  \sum_{b \in \mathcal{B}_i^{(t)}} \widehat{g}_b^{(t)}$ has zero mean and $\mathrm{I}_{d^*}$ covariance. Thus, 
\begin{align}
   H_M &= h \left(\frac{1}{\sqrt{MB}} \sum_{i=1}^M  \sum_{b \in \mathcal{B}_i^{(t)}} \widehat{g}_b^{(t)}\middle|  \left\{S_{{N}}^{(k)} \right\}_{k\in[t-1]} \right) \nonumber \\
  & \overset{(a)}\leq \frac{1}{2} d^*\log\left(2\pi e\right),
\end{align}
where $(a)$ follows  from the fact that for a fixed first and second moment, Gaussian distribution maximizes the entropy.

The distinction between the proof of the bound in Case 1 and Case 2 in Theorem 1 is  in the lower bound on the term $H_M$. We start by providing  the lower bound that is  used for proving  Case 1.

\subsection{Lower bound on \texorpdfstring{$H_M$}{H\textunderscore M} for Case 1 in Theorem 1} 
For the lower bound, we will rely heavily on the assumption  that the elements of $\widehat{g}_b^{(t)}$ are independent and the interesting result that gives Berry-Esseen style bounds for the entropic central limit theorem \cite{bobkov2014berry}. In particular, in its simplest form, the result states that for IID zero mean random variables $X_i$, the entropy of the normalized sum $T_m = \frac{1}{\sqrt{M}} \sum_{i=1}^M X_i$ approaches the entropy of a Gaussian random variable $\Phi_{\sigma^2}$ with the same variance $\sigma^2$ as $X_i$, such that the following is always satisfied
\begin{equation}\label{eq:berry_esseen_entropy}
h(\Phi_{\sigma^2}) - h(T_M) \leq \tilde{C} \frac{\mathbb{E}|X_i|^4}{M},
\end{equation}

Using~\eqref{eq:berry_esseen_entropy}, we can find a lower bound for $H_M$ as follows:
\begin{align}
H_M 
&= h \left(\frac{1}{\sqrt{MB}} \sum_{i=1}^M  \sum_{b \in \mathcal{B}_i^{(t)}} \widehat{g}_b^{(t)}\middle|  \left\{S_{{N}}^{(k)} \right\}_{k\in[t-1]} \right)  \nonumber \\
&= \sum_{j=1}^{d^*} h \underbrace{\left( \frac{1}{\sqrt{MB}}  \sum_{b \in \mathcal{B}_i^{(t)}} \sum_{i=1}^M \widehat{g}_b^{(t)}[j]\middle |   \left\{S_{{N}}^{(k)} \right\}_{k\in[t-1]}\right)}_{\text{variance = $1$}} \nonumber \\
&\stackrel{\eqref{eq:berry_esseen_entropy}}\geq\!\! \sum_{j=1}^{d^{*}}\! \left(\! h(\Phi_{1}) - \frac{C_{0,\bar{g}}}{B}\!\right) {=} \frac{d^*}{2} \log\left(2\pi e\right) {-} \frac{C_{0,\bar{g}}}{MB}.
\end{align}
In other words, we have the following bound on $H_M$
\begin{equation}\label{eq:bound_H_M}
  \frac{d^*}{2} \log\left(2\pi e\right) - \frac{d^{*}C_{0,\bar{g}}}{MB} \leq  H_M \leq \frac{d^*}{2} \log\left(2\pi e\right).  
\end{equation}

By substituting~\eqref{eq:bound_H_M} in~\eqref{eq:mutual_info_restated} (lower bound for $M=N-1$ and upper bound for $M=N$), we get that
\begin{align}
    I\left(\mathbf{x}^{(t)}_N ;   S_{{N}}^{(t)}\middle|  \left\{S_{{N}}^{(k)} \right\}_{k\in[t-1]} \right)
    &= H_N - H_{N-1} + d^*\log\left(\frac{N}{N-1}\right) \nonumber \\
    &\leq \frac{d^*}{2}\log\left(\frac{N}{N-1}\right) + \frac{d^*C_{0,\bar{g}}}{(N-1)B}.
\end{align}
This concludes the proof of  Case 1.

\subsection{Lower bound on \texorpdfstring{$H_M$}{H\textunderscore M} for Case 2 in Theorem 1} 
The proof of this  lower bound  relies on the   entropic central limit theorem for the vector case \cite{eldan2020clt} and Lemma 1 which will be stated later in this section. We start by giving the entropic central limit theorem for the case of IID random vectos~\cite{eldan2020clt}. 

\begin{thm}[Entropic central limit theorem \cite{eldan2020clt}]
Let $\mathbf{q}$ be a $\sigma$-uniformly log concave  $d$-dimensional random vector  with $\mathbb{E}[\mathbf{q}] = 0$ and non-singular covariance matrix $\Sigma$.  Additionally, let $\mathbf{z} \sim \mathcal{N}(0, \Sigma)$ be a Gaussian vector  with the same covariance as $\mathbf{q}$, and let $\gamma \sim \mathcal{N}(0, \mathrm{I}_{d}) $ to be a standard Gaussian. The entropy of the normalized sum $T_M = \frac{1}{\sqrt{M}} \sum_{i=1}^M \mathbf{q}_i$, where $\mathbf{q}_i$'s are random samples,  approaches the entropy of a Gaussian random vector $Z$, such that the following is always satisfied 

\begin{equation}\label{eq:entropy_vector}
    \text{Ent}(T_M||\mathbf{z})\leq \frac{2(d+2(\text{Ent}(\sqrt{\sigma} \mathbf{q}||\gamma)}{M\sigma^4}, 
\end{equation}
where $\text{Ent}(T_M||\mathbf{z})$ is the relative entropy. 
\end{thm}
\begin{lemma}\label{lemma:entropy}
Given a   random vector  $\mathbf{q} \in \mbb{R}^d$ with a distribution $f_{\mathbf{q}} (y)$ and   Cov$({\mathbf{q}}) = \Sigma$, and defining    $\mathbf{z} \sim \mathcal{N}(0, \Sigma)$ to  be a Gaussian vector  with the same covariance as $\mathbf{q}$,  for $\sigma >0$ , we get  
\begin{align}
 \text{Ent}(\sqrt{\sigma }\mathbf{q}||\mathbf{z}) &= - h(\mathbf{q}) -\frac{d}{2} \log(\sigma)+ \frac{d}{2} \log(2 \pi)\nonumber \\& + \frac{1}{2} \log(| \Sigma|) +  \sigma\frac{d}{2} \\ 
    \text{Ent}(\mathbf{q}||\mathbf{z}) &= h(\mathbf{z}) - h(\mathbf{q}) \label{eq-entropy2} 
\end{align}
\end{lemma}
%In the following, we state this resultthat for $\sigma$-log concave    $d$-dimensional random vector  $X$ with $\mathbb{E}[x] = 0$ and Cov$(X) = \Sigma$. \\

 Given the assumption that  $\widehat{g}_b^{(t)}$ has a  $\sigma$-log concave distribution  while both the term $\frac{1}{\sqrt{MB}} \sum_{i=1}^M  \sum_{b \in \mathcal{B}_i^{(t)}} \widehat{g}_b^{(t)}$  and $\widehat{g}_b^{(t)}$  have  an identity  covariance matrix $ \Sigma = \mathrm{I}_{d^*}$  given $\left\{S_{{N}}^{(k)} \right\}_{k\in[t-1]}$, we can  use ~\eqref{eq:entropy_vector} with  $\mathbf{z} \sim \mathcal{N}(0, \mathrm{I}_{d^*}) $. Furthermore, by using Lemma \ref{lemma:entropy}, we get  \begin{equation}\label{d}
    h(\mathbf{z})- H_M \leq \frac{d^* C_{1,\bar{g}} - C_{2,\bar{g}}}{MB}, 
\end{equation}
where, $C_{1,\bar{g}} = \frac{2\left( 1+\sigma +\log(2\pi) - \log(\sigma) \right)}{\sigma^4} $ and $C_{2,\bar{g}} = \frac{4h\left(\widehat{g}(b, \bm {\theta}^{(t)})\right)}{\sigma^4}$, and     $h(\widehat{g}(b, \bm {\theta}^{(t)}))$  is the entropy  of  the  random vector $\bar{g}_i(b, \bm {\theta}^{(t)})$  after whitening.

Finally, using the fact that   the entropy of  the Gaussian random vector $\mathbf{z}$ with  covariance $\mathrm{I}_{d^*}$ is given by $h(\mathbf{z}) = \frac{d^*}{2}\log(2 \pi e)$, 
we get the following bound on $H_M$
\begin{equation}\label{eq:bound_H_M1}
  \frac{d^*}{2} \log\left(2\pi e\right) - \frac{d^* C_{1,\bar{g}} - C_{2,\bar{g}}}{(N-1)B}  \leq  H_M \leq \frac{d^*}{2} \log\left(2\pi e\right).  
\end{equation}

By substituting~\eqref{eq:bound_H_M1} in~\eqref{eq:mutual_info_restated} (lower bound for $M=N-1$ and upper bound for $M=N$), we can now upper bound the mutual information term as follows
\begin{align}
      &I\left(\mathbf{x}^{(t)}_N ;   S_{{N}}^{(t)}\middle|  \left\{S_{{N}}^{(k)} \right\}_{k\in[t-1]} \right)\nonumber \\
    &= H_N - H_{N-1} + \frac{d^*}{2}\log\left(\frac{N}{N-1}\right) \nonumber \\
    &\leq \frac{d^*}{2}\log\left(\frac{N}{N-1}\right) +  \frac{d^* C_{1,\bar{g}} - C_{2,\bar{g}}}{(N-1)B}.
\end{align}
This concludes the proof of Theorem~1.

\section{Proof of Corollary 1}\label{append:proof_Corollary}
In the following, we define $S_{{N}}^{(t)} = \frac{1}{{N}}\sum_{i = 1}^N \mathbf{x}_i^{(t)}$. Using this notation, we can upper bound $I_{\rm priv/data}$ as follows
\begin{align}\label{co1}
     &I_{\rm priv/data} 
    %  = {I}\left(\mcal{D}_i ;  \left \{ \frac{1}{N}  \sum_{i\in \mathcal{N}} \mathbf{x}^{(k)}_i  \right\}_{k \in [T]} \right) \nonumber \\
     = {I}\left(\mcal{D}_i ;  \left \{ S_{{N}}^{(k)}  \right\}_{k \in [T]} \right) \nonumber \\
    &\overset{(a)}= \sum_{t =1}^{T} {I}\left(\mcal{D}_i ; S_{{N}}^{(t)}\middle|  \left \{ S_{{N}}^{(k)}  \right\}_{k \in [t-1]} \right)  \nonumber \\
    & \overset{(b)}\leq \sum_{t =1}^{T} {I}\left(\mcal{B}_i^{(t)} ;S_{{N}}^{(t)} \middle| \left \{ S_{{N}}^{(k)}  \right\}_{k \in [t-1]} \right)  \nonumber \\
    &\overset{(c)}\leq   \sum_{t =1}^{T}\! \underbrace{ I\!\left( \mathbf{x}_i^{(t)} \!\left(\mcal{B}_i^{(t)} ; \left \{\! S_{{N}}^{(k)}  \!\right\}_{\!\!k \in [t-1]}\!\right); S_{{N}}^{(t)} \middle| \left \{ \!S_{{N}}^{(k)}  \!\right\}_{k \in [t-1]} \!\right) }_{\text{This is bounded by the result in Theorem 1}}\!. 
\end{align}
where: (a) comes from the chain-rule; (b) from data processing inequality $D_i \rightarrow B_i^{(t)} \rightarrow \mathbf{x}_i^{(t)}$, where $B_i^{(t)}$ is the sampled mini-batch from the data set of node $i$;  (c) from data processing inequality $ B_i^{(t)} \rightarrow \mathbf{x}_i^{(t)} \rightarrow \frac{1}{N}  \sum_{i\in \mathcal{N}} \mathbf{x}^{(t)}_i  $;. Combining the results given in  the two cases of Theorem 1 with \eqref{co1} concludes the proof of Corollary 1.

\balance
\section{Proof of Lemma 1 }

\begin{align}\label{eq: lemma proof}
    &\text{Ent}(\sqrt{\sigma }\mathbf{q}||\mathbf{Z}) =  \text{Ent}(\mathbf{q}'||\mathbf{Z}) = \int f_{ \mathbf{q}'} (y) \log \frac{f_{\mathbf{q}'} (y)}{f_{\mathbf{Z}} (y)} dy \nonumber \\
    &=  \int f_{\mathbf{q}'} (y) \log f_{\mathbf{q}'} dy  -   \int f_{\mathbf{q}'}(y)  \log f_{\mathbf{Z}} (y) dy \nonumber  \\ 
    & \overset{(a)}= - h(\mathbf{q}') + \frac{d}{2} \log(2 \pi) \nonumber \\
    &\qquad + \frac{1}{2} \log(| \Sigma|)  + \frac{1}{2} \int f_{\mathbf{q}'}(y) y^T \Sigma^{-1}  y dy 
    \nonumber \\  & \overset{(b)}= - h(\mathbf{q}) -\frac{d}{2} \log(\sigma)+  \frac{d}{2} \log(2 \pi) \nonumber \\
    &\quad + \frac{1}{2} \log(| \Sigma|) + \frac{1}{2} \int f_{\mathbf{q}'}(y)  \text{ Tr}(\Sigma^{-1} y^T  y ) dy \nonumber \\  
    & \overset{(c)}= - h(\mathbf{q}) -\frac{d}{2} \log(\sigma)+ \frac{d}{2} \log(2 \pi) \nonumber \\
    &\qquad + \frac{1}{2} \log(| \Sigma|) + \frac{1}{2}  \text{ Tr}   \left( \Sigma^{-1} \int     f_{\mathbf{q}'}(y)  y^T  y  dy \right) \nonumber \\  
    & = - h(\mathbf{q}) -\frac{d}{2} \log(\sigma)+ \frac{d}{2} \log(2 \pi)\nonumber \\
    &\qquad + \frac{1}{2} \log(| \Sigma|) + \frac{1}{2}  \text{ Tr}  \left(  \Sigma^{-1}  \mathbb{E}_{\mathbf{q}'} [ \mathbf{q}'^{T} \mathbf{q}']  \right)
      \nonumber\\  
      & \overset{(d)}= - h(\mathbf{q}) + \frac{d}{2} \log(\frac{2 \pi}{\sigma}) + \frac{1}{2} \log(| \Sigma|) + \frac{1}{2} \sigma   \text{ Tr}  \left(  \Sigma^{-1}   \Sigma   \right)
    \nonumber \\  
    & = - h(\mathbf{q}) + \frac{d}{2} \log(\frac{2 \pi}{\sigma}) + \frac{1}{2} \log(| \Sigma|) + \sigma\frac{d}{2},
\end{align}
where: $\rm Tr$ represents the trace function; $(a) $ follows from using the multivariate distribution of the Gaussian vector $\mathbf{z}$; $(b)$  using the scaling property of the entropy with $\mathbf{q}' = \sqrt{\sigma} \mathbf{q}$; $(c)$ from follows from using the linearity of the trace function; finally $(d)$  from using the linear transformation of the random vector $\mathbf{q}' = \sqrt{\sigma} \mathbf{q}$ and the fact that $\mathbf{q}$ has the same covariance matrix $\Sigma$ as $\mathbf{z}$. 

The proof of \eqref{eq-entropy2} follows directly by substituting $\sigma =1 $ in  the equation  \eqref{eq: lemma proof} and using entropy  of a Gaussian vector with covariance $\Sigma$.  

\section{Overview of MINE} 
\label{subsec:mine}
In our empirical evaluation in Section~\ref{sec:eval}, we use the Mutual Information Neural Estimator (MINE)~\cite{belghazi2018mine} to estimate the mutual information, which is the state-of-the-art method for mutual information estimation \cite{belghazi2018mine}. Specifically, given random vectors $X$ and $Z$, and a function family parameterized by a neural network $\mathcal{F}=\{T_{\theta}:X\times Z\rightarrow\mathbb{R}\}_{\theta\in\Theta}$, the following bound holds:
\begin{equation}
    I(X;Z)\geq I_{\Theta}(X;Z),
\end{equation}
where $I_{\Theta}(X;Z)$ is the neural mutual information measure defined as:
\begin{equation}
    I_{\Theta}(X;Z)=\sup_{\theta\in\Theta}\mathbb{E}_{\mathbb{P}_{XZ}}[T_{\theta}]-\log(\mathbb{E}_{\mathbb{P}_{X}\otimes\mathbb{P}_{Z}}[e^{T_{\theta}}]),
\end{equation}
$\mathbb{P}_{X}$ and $\mathbb{P}_{Z}$ are the marginal distribution of $X$ and $Z$ respectively, $\mathbb{P}_{XZ}$ is the joint distribution of $X$ and $Z$, and $\mathbb{P}_{X}\otimes\mathbb{P}_{Z}$ is the product of marginals $\mathbb{P}_{X}$ and $\mathbb{P}_{Z}$. As an empirical estimation of $I_{\Theta}(X;Z)$, MINE is implemented as 
\begin{equation}
\label{eq:mine}
    \widehat{I(X;Z)}_K=\sup_{\theta\in\Theta}\mathbb{E}_{\mathbb{P}_{XZ}^{(K)}}[T_{\theta}]-\log(\mathbb{E}_{\mathbb{P}_{X}^{(K)}\otimes\mathbb{P}_{Z}^{(K)}}[e^{T_{\theta}}]),
\end{equation}
where $\mathbb{P}_{(\cdot)}^{(K)}$ is the empirical distribution of $\mathbb{P}_{(\cdot)}$ with $K$ IID samples. Finally, solving Eq. \ref{eq:mine} (i.e. get the MI estimation) can be achieved by solving the following optimization problem via gradient ascent:
\begin{align*}
% \label{eq:mine_opt}
    &\widehat{I(X;Z)}_K \nonumber \\
    &= \max_{\theta\in\Theta}\left\{\!\frac{1}{K}\!\sum_{k=1}^{K}T_\theta(x_k,z_k)
    -\log\left(\frac{1}{K}\sum_{k=1}^{K}e^{T_\theta(x_k,\bar{z}_k)}\right)\right\},
\end{align*}
where $(x_k,z_k)$ is the $k$-th sample from $\mathbb{P}_{XZ}$ and $\bar{z}_k$ is the $k$-th sample from $\mathbb{P}_{Z}$.

\end{document}